%% file: main.tex
\documentclass[11pt]{article}

\usepackage[preprint]{acl}

\usepackage{times}
\usepackage{latexsym}

\usepackage[T1]{fontenc}

\usepackage[utf8]{inputenc}

\usepackage{microtype}
\usepackage{inconsolata}

\usepackage{graphicx}
\usepackage{booktabs}
\usepackage{multirow}
\usepackage{amsmath}
\usepackage{amssymb}
\usepackage{xcolor}
\usepackage{colortbl}
\usepackage{tcolorbox}
\usepackage{array}
\usepackage{etoc}
\providecommand{\ours}{\textup{\textsc{Skill++}}}
\usepackage{tikz}
\usetikzlibrary{shapes.geometric,arrows.meta,calc}
\newcommand*{\circled}[1]{\tikz[baseline=(c.base)]\node[circle,draw,line width=0.4pt,inner sep=0pt,minimum size=1.05em,font=\footnotesize] (c) {#1};}
\usepackage{algorithm}
\usepackage{algpseudocode}

\title{\texorpdfstring{\ours{}}{Skill++}: Towards Lifelong Agents via Routing-Aware Skill Refinement}

\author{%
  Hanrong Zhang$^{1*\dagger}$ \quad Minghao Liang$^{2*}$ \quad Jiawei Wu$^{3}$ \quad
  Yankai Chen$^{4,5\dagger}$ \quad Shicheng Fan$^{1}$ \\
  \bfseries Zhiwei Liu$^{6}$ \quad Enze Ma$^{1}$ \quad Junwei Yuan$^{7}$ \quad
  Tianyi Wang$^{7}$ \quad Dehai Min$^{1}$ \\
  \bfseries Kening Zheng$^{1}$ \quad Xue Liu$^{4,5}$ \quad Philip S. Yu$^{1\dagger}$}

\begin{document}
\maketitle
\makeatletter
{\long\def\@makefntext#1{\noindent\raggedright#1}\footnotetext{%
  \begin{tabular}{@{}r@{\,}l@{\hspace{1.5em}}r@{\,}l@{}}
    $^{1}$ & University of Illinois Chicago & $^{2}$ & Trip.com Group \\
    $^{3}$ & Rutgers University & $^{4}$ & MBZUAI \\
    $^{5}$ & McGill University & $^{6}$ & Recursive \\
    $^{7}$ & Independent Researcher & & \\[3pt]
    $^{*}$ & \multicolumn{3}{@{}l@{}}{Equal contribution.} \\
    $^{\dagger}$ & \multicolumn{3}{@{}l@{}}{Corresponding authors: \texttt{yankaichen@acm.org},} \\
    & \multicolumn{3}{@{}l@{}}{\texttt{\{hzhan135, psyu\}@uic.edu}}
  \end{tabular}}}
\makeatother

\begin{abstract}
\input{sections/00_abstract}
\end{abstract}

\etocdepthtag.toc{mtmain}
\input{sections/01_introduction}
\input{sections/02_related_work}
\input{sections/03_method}
\input{sections/04_experiments}

\input{sections/05_conclusion}

\section*{Limitations}
Our study is limited to text. Every skill consists of a name, a description, and a textual body, and both the queries and the router's input are text. Skills or queries involving images, audio, or other modalities may interfere in ways that textual metadata cannot capture. We leave multimodal skills and queries to future work.

\bibliography{refs}

\appendix
\clearpage
\setcounter{figure}{0}
\setcounter{table}{0}
\renewcommand{\thefigure}{A\arabic{figure}}
\renewcommand{\thetable}{A\arabic{table}}
\renewcommand{\theHfigure}{A\arabic{figure}}
\renewcommand{\theHtable}{A\arabic{table}}
\etocdepthtag.toc{mtappendix}
\etocsettagdepth{mtmain}{none}
\etocsettagdepth{mtappendix}{subsection}
\etocsettocstyle{\section*{Appendix Contents}}{}
\tableofcontents

Appendix~\ref{app:impl} details the implementation of \ours{}, Appendix~\ref{app:exp_setup} describes the experimental setup, and Appendix~\ref{app:exp_supp} gives evidence for each RQ.

\input{sections/A_implementation}
\input{sections/B_experiments_supplement}

\end{document}

%% file: sections/00_abstract.tex
Large language model (LLM) agents extend their capabilities with skills, each with a name, a description, and a body. Under progressive disclosure, an LLM router selects skills by name and description alone and loads only the selected bodies. Skill libraries grow continually as users and agents add new skills. Because the router relies on this metadata, a similar new skill can displace an earlier one, breaking previously correct routing even though the earlier skill and the router are unchanged. This routing interference is aggravated for public skills, whose descriptions may not match their bodies. We formulate the problem as lifelong skill routing, a class-incremental continual learning problem. A solution must keep earlier skills selectable without suppressing new ones. Existing approaches fall short, since continually retraining the router risks catastrophic forgetting and rewriting methods either ignore similar skills or require per-skill feedback. We therefore propose \ours{}, which rewrites only the name and description of each incoming skill, without router training or labeled data. The new description states what the body supports, how the skill differs from its closest predecessor, and what related work it excludes. As a library grows to 731 skills over ten sessions, \ours{} selects all required skills for 76.3\% of queries, versus 36.4\% under the original metadata. Across three benchmarks and five LLM routers, it outperforms all compared approaches, and it also improves on the original metadata in the Codex CLI agent harness.

%% file: sections/01_introduction.tex
\section{Introduction}
\label{sec:intro}

\begin{figure}[t]
  \centering
  \includegraphics[width=\columnwidth]{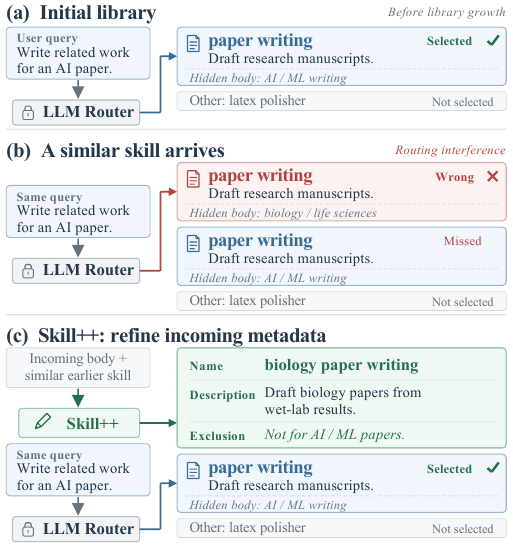}
  \caption{This figure presents an example of routing interference, to show why incoming metadata needs refinement. The router reads only skill names and descriptions. (a)~For a query about an AI paper, it selects the AI paper-writing skill. (b)~A biology paper-writing skill arrives with the same metadata, and the router selects it instead. (c)~\ours{} refines only the incoming skill against the earlier one, so the router selects the correct skill again.}
  \label{fig:example}
\end{figure}

\begin{figure*}[t]
  \centering
  \includegraphics[width=\textwidth]{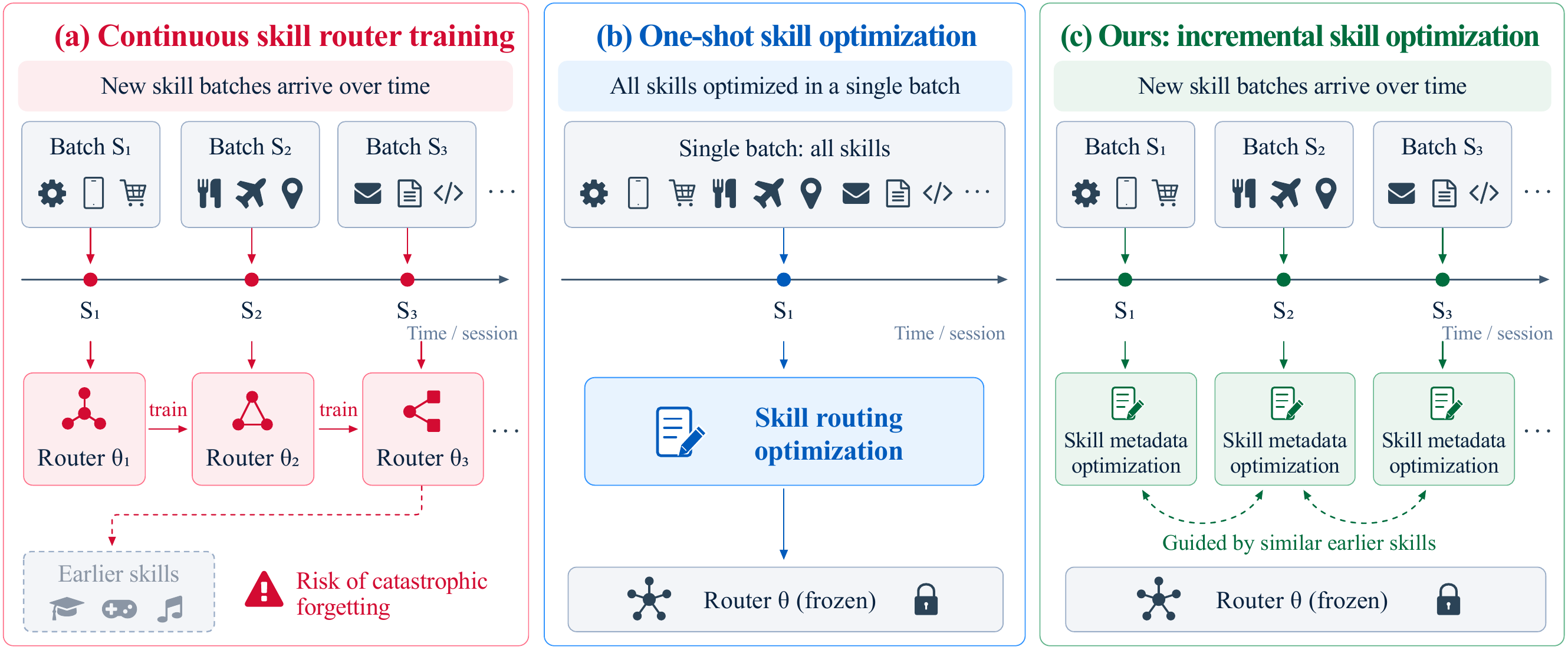}
  \caption{This figure compares three ways of handling a growing skill library, to position our problem. (a)~Continuous router training follows arriving skills but retrains the router at every step, risking catastrophic forgetting of earlier skills. (b)~One-shot skill optimization refines a fixed skill set once, so it ignores skills that arrive later. (c)~\ours{} keeps the router frozen and refines each new batch against similar earlier skills, so both earlier and new skills are selected for the queries that need them.}
  \label{fig:settings}
\end{figure*}

LLM agents \citep{zou2026humanagent,zou2026interruptible} increasingly acquire new capabilities through skills \citep{song2026drclaw}. A skill is a reusable package that teaches an agent how to perform a specific type of task. It consists of a name, a short description, and a body that holds instructions, code, and reference material, such as a guide to drawing charts with matplotlib. Because loading every body would exceed the context window, agents follow a design known as progressive disclosure, in which a skill's body is read only after the skill is selected \citep{zheng2026skillrouter}. Skill selection therefore relies on a router that sees only the names and short descriptions of skills, which we call routing metadata. As a result, \textbf{a skill is effective only if the router selects it when a query calls for it}. Such selection would be straightforward in a small, fixed library, but \textbf{in practice, skill libraries rarely stay fixed}. Users continually install community skills for new tools and domains, and some agents even generate new skills without human authoring \citep{zhang2026coevoskills,yan2026openskill,yang2026timeevo}. For instance, a recent analysis covers 138K public skill files \citep{zhang2026skillmd}, and controlled audits \citep{kang2026skillseam} show that a router reading only names and descriptions easily confuses synonymous or overlapping skills (Fig.~\ref{fig:example}).

Because new skills continue to arrive while earlier ones must remain usable, routing in a growing library is naturally a continual learning problem (also known as lifelong learning). More specifically, it is class-incremental, because new kinds of skills arrive over time while the router must still recognize the earlier ones. We formulate it as lifelong skill routing (Fig.~\ref{fig:settings}c), with skills arriving in batches, the router kept frozen, and only the metadata of each incoming skill allowed to change. This setting raises two challenges. \circled{1}~Whether a skill's metadata identifies the skill \textbf{depends on the other skills in the library}. A description that singled out a skill at insertion can lose its distinctiveness once similar skills arrive. The problem is more severe for skills collected from public sources, whose quality varies widely and whose routing metadata is often defective \citep{zhang2026skillmd,hao2026skillinjection,zhang2025asb}, so a name or description may not reflect what the body covers. In our experiments with the original metadata, 62 of the 115 queries that were routed correctly when their required skills first became available fail once the library reaches 731 skills (Appendix~\ref{app:rq1_supp}). We call this effect routing interference, a form of forgetting at access rather than in parameters. It takes two forms. In \emph{substitution}, a similar incoming skill is selected in place of the required skill (Fig.~\ref{fig:example}a--b). In \emph{crowding}, incoming skills occupy the few selected slots and thereby exclude the required skill. These two forms cause 87.5\% of the errors under the original metadata (Fig.~\ref{fig:rq4-error}a).

\circled{2}~Avoiding this interference is itself difficult, because \textbf{protecting earlier skills conflicts with adding new ones}. This conflict mirrors the stability--plasticity dilemma of continual learning \citep{delange2022survey}, where shared parameters must retain earlier knowledge (stability) while acquiring new knowledge (plasticity). In our setting, the same tension arises in skill metadata. Stability requires that each earlier skill is still selected for the queries that need it after new skills arrive, and plasticity requires that each new skill is selected for its own queries. Concretely, a narrow description prevents a new skill from competing for the queries of earlier skills but leaves it rarely selected, whereas a broad description makes it selectable but diverts queries meant for earlier skills.

Existing approaches address these challenges only in part. Retraining a learned router \citep{zheng2026skillrouter,wang2026r3skill} after each addition requires new training data and computation, and sequential updates can overwrite what the router learned about earlier skills, a problem known as catastrophic forgetting \citep{murtaza2026synthetic}. Rewriting the text the router reads avoids retraining, but current rewriters fall short in our setting. Specifically, descriptions written from each tool's own documentation ignore which similar skills the tool competes with \citep{lu2025toolde,guo2026tracefree}. Rewriting from routing errors \citep{zhou2026rewrite} requires labeled queries for each new skill, which our setting does not provide. Contrasting skills within a cluster needs no feedback but has been tested only on fixed libraries, so its effect on earlier skills is unknown \citep{wang2026capability}. Moreover, none of them measures stability and plasticity together as the library grows (Fig.~\ref{fig:settings}).

To address both challenges, we propose \ours{}, a simple yet effective method that refines the metadata of each incoming skill against the earlier skills it is most likely to be confused with (Fig.~\ref{fig:example}c). Our key observation is that routing metadata cannot be written in isolation, because the router needs to know not only what a skill does but also what separates it from similar skills in the library. \ours{} realizes this idea with three components. Description differentiation rewrites the description from the skill's own body, which keeps it aligned with what the skill does, and centers it on what distinguishes the skill from its most similar predecessor. Scope exclusions state related work that the skill does not cover, and name clarification qualifies a name that overlaps with an earlier one. Differentiation and clarified names thus reduce substitution, and exclusions reduce crowding (challenge~\circled{1}, Fig.~\ref{fig:rq4-error}c). Earlier skills stay selectable while each new skill is selected for its own queries, which addresses the trade-off of challenge~\circled{2}. Refinement is performed once at insertion, without router training or labeled feedback, so routers use the revised metadata directly.

In summary, our contributions are as follows:
\begin{itemize}
    \item To the best of our knowledge, we are the first to formalize lifelong skill routing, in which skills arrive sequentially into the skill library. Within this formulation, we show that routing interference prevents earlier skills from being selected for the queries that need them, even though their content and the router are unchanged, and that mitigating it raises the stability--plasticity dilemma of traditional continual learning at the level of skill metadata rather than model parameters.
    \item We propose \ours{}, a simple yet effective method that incrementally optimizes the skill library for skill routing. Whenever a skill arrives, \ours{} refines only that skill's name and description against similar earlier skills, without router training or labeled feedback.
    \item We evaluate \ours{} on three skill benchmarks (SRA-Bench, SkillRet, and SkillRouter) with five LLM routers and the Codex CLI harness. Averaged over the five routers, \ours{} improves routing accuracy at the final session by 23.1 percentage points (pp) over the original metadata, and achieves state-of-the-art performance among all compared approaches across benchmarks and routers.
\end{itemize}

%% file: sections/02_related_work.tex
\section{Related Work}
\label{sec:related}

\paragraph{Skill routing.}
Prior work improves skill routing either by training the router or by rewriting the text it reads. Router-side methods train retrievers and rerankers to select individual skills \citep{kang2026skillret,zheng2026skillrouter}, and later ones model the compatibility among candidates to select complementary skill sets \citep{wang2026r3skill}. These methods \textbf{leave descriptions unchanged}, although audits show that overlapping skills and weak descriptions cause routing failures \citep{kang2026skillseam,zhang2026skillmd}. Text-side methods rewrite the descriptions instead. Descriptions derived from each tool's own documentation \citep{lu2025toolde,guo2026tracefree} \textbf{never see the similar tools in the library}, and enriching them yields little benefit on a production catalog \citep{gillespie2026enterprise}. Later methods make these differences explicit, either by learning from execution traces or labeled routing errors that \textbf{must be collected for each skill} \citep{lu2026docschisel,zhou2026rewrite,ghoshal2026jtpro} or by contrasting the skills within a cluster and \textbf{recompiling the cluster whenever a skill is added} \citep{wang2026capability}. \ours{} overcomes both limitations by refining only the incoming skill at insertion, with a description grounded in its body. Because the router stays unchanged, it also complements router-side methods.

\paragraph{Continual learning.}
Our setting is a class-incremental continual learning problem, because earlier skills must remain selectable as new skills arrive. Classic methods retain earlier tasks by constraining parameter updates \citep{kirkpatrick2017overcoming,lopezpaz2017gem,delange2022survey}. In the class-incremental scenario, the hardest of the three standard ones, confusion between old and new classes is a main source of forgetting \citep{vandeven2022three,zhou2024classincremental,zhang2025classincremental,yao2026uncertainty}. Beyond parameters, LLM agents can also store capabilities outside the model in a growing skill library \citep{wang2024voyager,huang2026memorysurvey,wu2026gam,zhu2026rsiagent}. \textbf{Forgetting then resurfaces at retrieval time}, because new entries or retriever fine-tuning make earlier entries harder to retrieve \citep{hu2026memorycl,murtaza2026synthetic,song2026shadowing,chen2026scaling}. Existing remedies \textbf{retrain the retriever or add structures} such as skill graphs \citep{murtaza2026synthetic,bai2026skilldag}. \ours{} instead keeps the router frozen and edits only the metadata of the incoming skill. Unlike standard class-incremental learning, the class descriptions change rather than the classifier, and a new skill may join an old class. We therefore evaluate stability and plasticity separately.

%% file: sections/03_method.tex
\section{Method}
\label{sec:method}

\begin{figure*}[t]
  \centering
  \includegraphics[width=\textwidth]{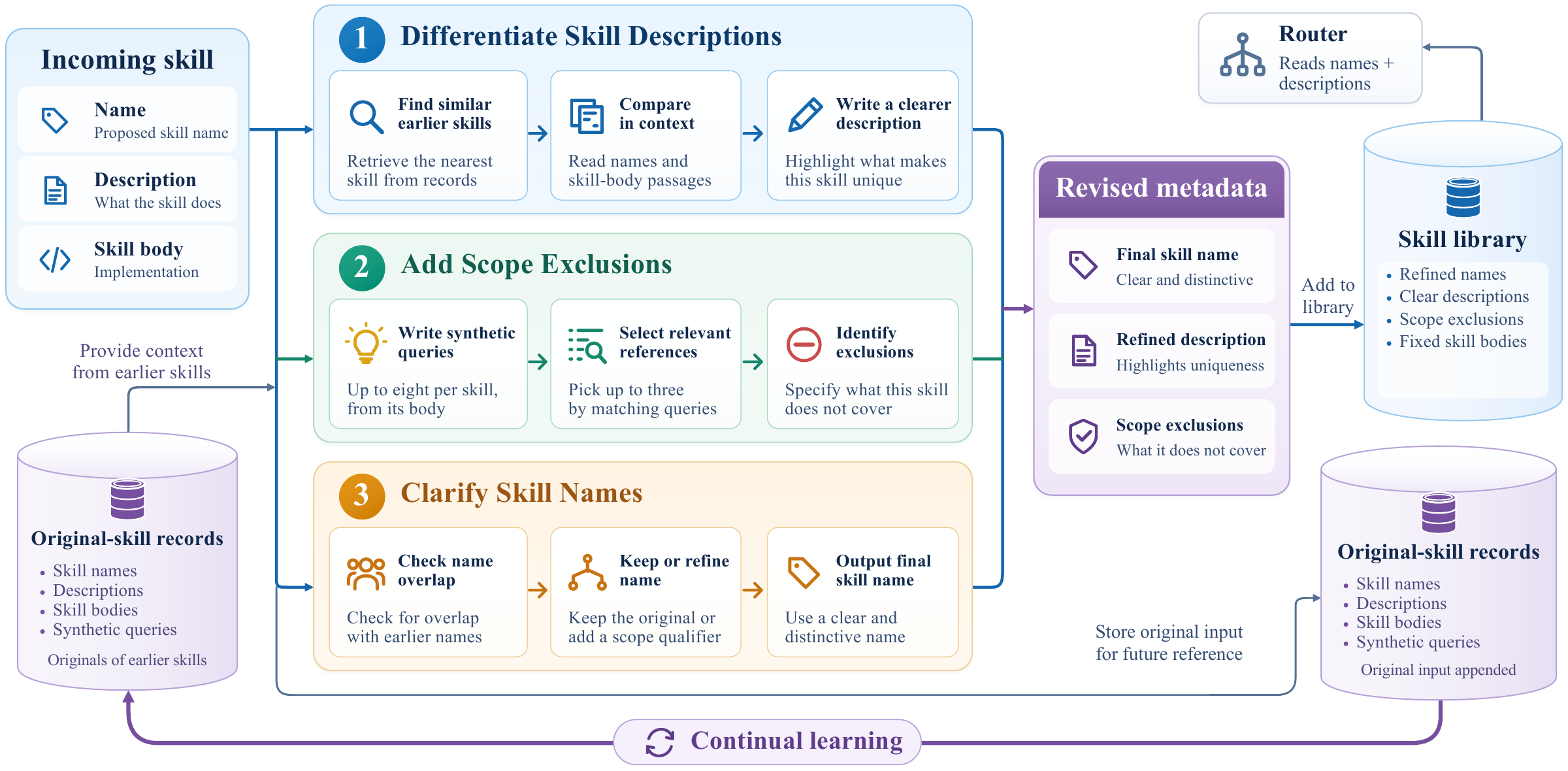}
  \caption{This figure gives an overview of \ours{}, to show how it refines an incoming skill before the router reads it. \ours{} first retrieves similar earlier skills from a record of their original names, descriptions, bodies and synthetic queries. Three components then refine the new skill independently. Description differentiation rewrites its description around what makes it unique, scope exclusions list related work that it does not cover, and name clarification makes an overlapping name more specific. The refined name and description enter the library that the router reads, and the original skill joins the record for future additions.}
  \label{fig:overview}
\end{figure*}

\subsection{Problem Formulation}
\label{sec:formulation}

\paragraph{Continual learning setting.}
We formulate skill routing in a growing library as class-incremental continual learning. In standard class-incremental learning~\citep{vandeven2022three,zhou2024classincremental}, a model $f_\theta$ receives a sequence of tasks whose data $\mathcal{D}_1,\dots,\mathcal{D}_T$ introduce disjoint sets of classes. After each task, a learning algorithm $\mathcal{A}$ updates the parameters, and the model is tested on the pooled data $\mathcal{D}_{\le t}$ of all tasks so far (Fig.~\ref{fig:cil}a):
\begin{equation}
  \begin{aligned}
    \theta_t&=\mathcal{A}(\theta_{t-1},\mathcal{D}_t),\\
    \mathrm{Acc}_t&=\Pr_{(x,y)\sim\mathcal{D}_{\le t}}\bigl[f_{\theta_t}(x)=y\bigr].
  \end{aligned}
  \label{eq:cil}
\end{equation}
In our setting (Fig.~\ref{fig:cil}b), skills arrive in batches and are added one at a time, so step $t$ brings a single skill $c_t=(r_t,b_t)$. Its routing metadata $r_t=(n_t,d_t)$ consists of a name and a short description, and its body $b_t$ holds instructions and supporting material. Skills that serve the same kind of work form a skill class, such as the skills for plotting with matplotlib. Unlike in standard class-incremental learning, $c_t$ may join an existing class and then compete with its earlier members for the same queries. A frozen router reads the names and descriptions in a library $\mathcal{L}$ and returns the identifiers of $K$ skills, $\mathrm{top}_K(q;\mathcal{L},\xi)$, for a query $q$, where $\xi$ captures its randomness. Routing succeeds when the selection contains the annotated skills $S^\star(q)$, one for each class whose work $q$ needs. When $c_t$ arrives, a method observes $c_t$, the current library $\mathcal{L}_{t-1}$, and the records $\mathcal{H}_{t-1}$ of original metadata and bodies, but no queries, routing outcomes, or class labels. It may edit only $r_t$, which keeps the metadata of earlier skills stable. The edited metadata $\tilde r_t$ then updates the library and records as
\begin{equation}
  \mathcal{L}_t=\mathcal{L}_{t-1}\cup\{(\tilde r_t,b_t)\},\quad \mathcal{H}_t=\mathcal{H}_{t-1}\cup\{c_t\},
  \label{eq:library-update}
\end{equation}
where $\mathcal{L}_0$ and $\mathcal{H}_0$ hold the initial skills.

\paragraph{Objective.}
Queries follow a fixed deployment distribution $\mathcal{Q}$. Let $\mathcal{C}_t$ denote the skills that have arrived by step $t$, so the queries served at step $t$ follow $\mathcal{Q}_t(q)\propto\mathcal{Q}(q)\,\mathbf{1}[S^\star(q)\subseteq\mathcal{C}_t]$. The success rate after step $t$ is
\begin{equation}
  A_t=\Pr_{q\sim\mathcal{Q}_t,\,\xi}\!\left[S^\star(q)\subseteq\mathrm{top}_K(q;\mathcal{L}_t,\xi)\right].
  \label{eq:reward}
\end{equation}
It splits by whether a query needs the new skill:
\begin{equation}
  A_t=\omega_t\,A_t^{\mathrm{old}}+(1-\omega_t)\,A_t^{\mathrm{new}},
  \label{eq:split}
\end{equation}
where $\omega_t=\Pr_{q\sim\mathcal{Q}_t}[c_t\notin S^\star(q)]$, and $A_t^{\mathrm{old}}$ and $A_t^{\mathrm{new}}$ are the success rates of Eq.~\eqref{eq:reward} conditioned on $c_t\notin S^\star(q)$ and on $c_t\in S^\star(q)$. $A_t^{\mathrm{old}}$ measures stability and $A_t^{\mathrm{new}}$ measures plasticity. The two can conflict, because making $c_t$ more prominent can raise $A_t^{\mathrm{new}}$ while lowering $A_t^{\mathrm{old}}$. We seek an editing policy $\pi$ that maps each observation to an edit and maximizes
\begin{equation}
  \mathbb{E}_{\pi}\!\left[\frac{1}{T}\sum_{t=1}^{T} A_t\right],
  \label{eq:objective}
\end{equation}
where the expectation is taken over addition orders and metadata generation. Eqs.~\eqref{eq:library-update}--\eqref{eq:objective} instantiate Eq.~\eqref{eq:cil}, with the library $\mathcal{L}_t$ in the role of $\theta_t$, the policy $\pi$ in that of $\mathcal{A}$, and $A_t$ in that of $\mathrm{Acc}_t$.

\subsection{\texorpdfstring{\ours{}: Routing-Aware Refinement}{Skill++: Routing-Aware Refinement}}
\label{sec:refine}

\ours{} targets Eq.~\eqref{eq:objective} through a surrogate objective that uses only the information available when a skill is added, namely the incoming skill and the earlier skills. Routing errors concentrate among skills with similar metadata \citep{chen2026scaling}, so the incoming metadata should separate the skill from the earlier skills $\mathcal{N}_t$ it is most likely to be confused with, while staying faithful to its own body (Sec.~\ref{sec:intro}):
\begin{equation}
  \max_{r}\,\Gamma(r;\mathcal{N}_t)\ \ \text{s.t.}\ \ r\models b_t,\ |r|\le\ell_{\max}.
  \label{eq:surrogate}
\end{equation}
Here $\Gamma$ measures how well $r$ separates $c_t$ from the skills in $\mathcal{N}_t$, the fidelity constraint $r\models b_t$ requires every capability claimed in $r$ to be supported by $b_t$, and $\ell_{\max}$ bounds the length. Rather than searching over $r$ numerically, \ours{} realizes each term of Eq.~\eqref{eq:surrogate} with one part of its design. Description differentiation and name clarification separate $c_t$ from its most similar earlier skill $g_t$, and scope exclusions separate it from the related skills $\mathcal{G}_t$, so $\mathcal{N}_t=\{g_t\}\cup\mathcal{G}_t$. Grounding the capability description in $b_t$ targets fidelity, and dropping exclusion phrases from an overlong description enforces the length bound.

The three components (Fig.~\ref{fig:overview}) read the original records $\mathcal{H}_{t-1}$, so earlier revisions never affect later references. We write $\mathcal{M}_D$, $\mathcal{M}_E$, and $\mathcal{M}_N$ for the prompted LLM calls that write descriptions, exclusion phrases, and names, and Appendix~\ref{app:impl} gives their settings and summarizes one addition step in Algorithm~\ref{alg:skillpp}.

\begin{figure*}[t]
\centering
\includegraphics[width=\textwidth]{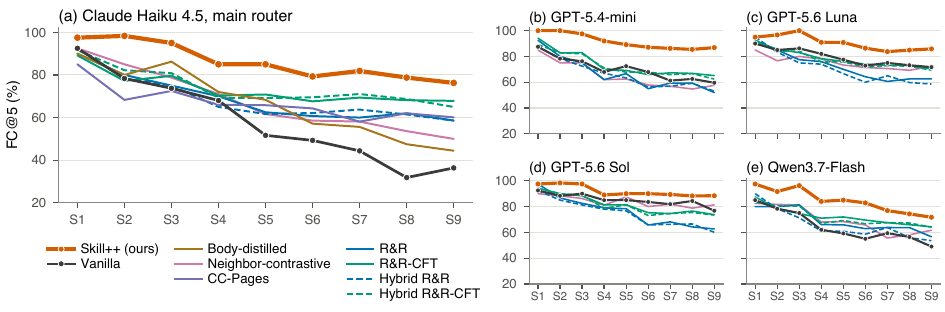}
\caption{This figure shows FC@5 as the library grows (sizes in Tab.~\ref{tab:sessions}), to test whether routing stays accurate. (a)~The main router, Claude Haiku 4.5 (Tab.~\ref{tab:rq1_traj}). (b)--(e)~Four other routers (Tab.~\ref{tab:routers}). With every router, Vanilla declines while \ours{} leads or ties at every checkpoint.}
\label{fig:routers}
\end{figure*}

\paragraph{Differentiating Skill Descriptions.}
This component writes the positive part of the metadata. A description written from the body alone tends to repeat capabilities that many similar skills share, so it gives the router little evidence to separate them. Differentiation therefore contrasts the incoming skill with the earlier skill whose original metadata is most similar,
\begin{equation}
  g_t=\underset{s:\,c_s\in\mathcal{H}_{t-1}}{\arg\max}\;\phi(r_s)^{\top}\phi(r_t),
  \label{eq:reference}
\end{equation}
where $\phi(r)$ is the unit-normalized embedding of the name and description in $r$. The call $\mathcal{M}_D$ reads both names and short passages from both bodies and writes a capability description $\hat d_t$ that leads with what sets the incoming skill apart. An original description can also misstate its body. It may overclaim, advertising work that the body does not support, so that the skill takes the queries of earlier skills. It may instead underclaim, omitting work that the body covers, so that the skill misses its own queries. Every claim of $\hat d_t$ must therefore come from the incoming body, which aligns the description with what the skill does, and the reference is never named.

\paragraph{Adding Scope Exclusions.}
This component writes the negative part and targets crowding. It tells the router which nearby work the incoming skill does not cover, so that the skill leaves those queries to related skills. Related skills may serve the same queries under different wording, so their metadata need not look alike, and \ours{} therefore compares skills through synthetic queries. When a skill is installed, a generation model writes a set $\tilde Q_s$ of up to eight queries from its body alone, and the record stores them with the skill. These queries never draw on evaluation queries or labels, and they serve only this search. Each earlier skill $s$ then receives three scores. $\sigma_1(s)$ asks whether the incoming metadata matches a typical query of $s$, $\sigma_2(s)$ whether the two skills serve similar queries, and $\sigma_3(s)$ whether their original metadata look alike. The reference set and the exclusion clause are
\begin{equation}
  \begin{aligned}
    \mathcal{G}_t&=\bigl\{s:\ c_s\in\mathcal{H}_{t-1},\ \bar\rho(s)\le 3\bigr\},\\
    e_t&=\mathcal{M}_E\bigl(\{r_s\}_{s\in\mathcal{G}_t}\bigr),
  \end{aligned}
  \label{eq:exclusion}
\end{equation}
where $\bar\rho(s)$ is the fused rank of $s$ under the three scores, with 1 for the best candidate (Eq.~\eqref{eq:fusion} in Appendix~\ref{app:impl}). The call $\mathcal{M}_E$ summarizes the responsibility of each reference in a short phrase, and $e_t$ joins these phrases after \texttt{Does not cover}.

\paragraph{Clarifying Skill Names.}
Names are short and prominent in the router's input, so two skills with similar names can be confused even when their descriptions differ. This component therefore targets substitution through the name. Let $h(n)$ denote the name head of $n$, its first word (Appendix~\ref{app:impl}). A conflict is flagged by
\begin{equation}
  z_t=\mathbf{1}\bigl[\exists\, c_s\in\mathcal{H}_{t-1}:\ h(n_s)\sqsubseteq n_t\bigr],
  \label{eq:rename}
\end{equation}
where $\sqsubseteq$ denotes substring containment. If $z_t=1$, the call $\mathcal{M}_N$ reads $n_t$, a passage from $b_t$, and the reference name $n_{g_t}$, and proposes a name qualified by the topic, workflow, or type of guidance in the body. The final name $\tilde n_t$ is this proposal if it has 4 to 60 characters, and $n_t$ otherwise.

\paragraph{Assembling Outputs.}
The revised metadata is
\begin{equation}
  \tilde r_t=(\tilde n_t,\ \hat d_t\oplus e_t),
  \label{eq:assemble}
\end{equation}
where $\oplus$ appends the exclusion clause. If the result exceeds a length threshold $\ell_{\max}$, exclusion phrases are dropped from the end, so the capability description is never truncated.

%% file: sections/04_experiments.tex
\providecommand{\ours}{\textup{\textsc{Skill++}}}%
\definecolor{tkframe}{RGB}{22,78,60}
\definecolor{tkfill}{RGB}{232,243,238}
\definecolor{tbband}{HTML}{EEF2F6}
\definecolor{tbours}{HTML}{E6F0FA}
\providecommand{\tbbest}[1]{\textbf{#1}}
\providecommand{\tbsecond}[1]{\underline{#1}}
\providecommand{\tbours}[1]{\rowcolor{tbours}\textbf{#1}}
\providecommand{\tbband}[2]{\rowcolor{tbband}\multicolumn{#1}{c}{\textbf{#2}}}
\newlength{\tbgap}\setlength{\tbgap}{0.8em}
\newcolumntype{N}[1]{>{\setbox0=\hbox{#1}\hbox to \wd0\bgroup\hfil}c<{\unskip\egroup}}
\newcolumntype{G}[1]{>{\hspace{\tbgap}\setbox0=\hbox{#1}\hbox to \wd0\bgroup\hfil}c<{\unskip\egroup}}
\providecommand{\tbdfont}{\scriptsize\color{black!55}}
\newcolumntype{D}[2]{>{\hspace{\tbgap}\setbox2=\hbox{\tbdfont#2}\setbox0=\hbox{#1}\hbox to \wd0\bgroup\hfil}c<{\unskip\egroup}}
\providecommand{\tbd}[1]{\unskip\egroup\raisebox{-0.4ex}{\hbox to \wd2{\hfil\tbdfont#1}}\hbox\bgroup}
\providecommand{\tbhead}[1]{\multicolumn{1}{c}{#1}}
\providecommand{\tbgaphead}[1]{\multicolumn{1}{>{\hspace{\tbgap}}c}{#1}}
\providecommand{\tbstub}[1]{\multirow{2}{*}[-0.6ex]{#1}}
\newlength{\tbnatural}
\newcommand{\tbfit}[4]{%
  \settowidth{\tbnatural}{\setlength{\tabcolsep}{0pt}\begin{tabular}{#3}#4\end{tabular}}%
  \setlength{\tabcolsep}{\dimexpr(#1-\tbnatural)/\numexpr2*#2\relax\relax}%
  \ifdim\tabcolsep<1pt\setlength{\tabcolsep}{1pt}\fi
  \begin{tabular}{#3}#4\end{tabular}}
\providecommand{\takeawaybox}[3]{%
  \begin{tcolorbox}[colback=tkfill, colframe=tkframe, coltext=tkframe, coltitle=white,
    fonttitle=\bfseries, title={#1}, arc=1.2mm, boxrule=0.6pt, left=4pt, right=4pt,
    top=3pt, bottom=3pt, toptitle=2pt, bottomtitle=2pt, before skip=6pt, after skip=6pt]
  \textbf{#2} #3
  \end{tcolorbox}}
\newenvironment{expitems}{\begin{list}{\textbullet}{%
  \setlength{\leftmargin}{0.9em}\setlength{\labelwidth}{0.5em}\setlength{\labelsep}{0.4em}%
  \setlength{\itemsep}{1pt}\setlength{\parsep}{0pt}\setlength{\topsep}{2pt}\setlength{\partopsep}{0pt}}}%
  {\end{list}}

\section{Experiments}
\label{sec:exp}

We evaluate \ours{} with four research questions (RQs) on performance (RQ1), the source of its gains (RQ2), generalization (RQ3), and the contribution of each component (RQ4).

\subsection{Experimental Setup}
\label{sec:setup}

\paragraph{Benchmarks.}
We build on the multi-skill queries of three benchmarks, SRA-Bench~\citep{su2026sra}, SkillRet~\citep{kang2026skillret} and SkillRouter~\citep{zheng2026skillrouter}. Each originally evaluates routing over a fixed skill library with queries annotated by their required skills, and we turn each into a class-incremental version with the same recipe. Over ten sessions (S0 to S9), new skill classes arrive together with the queries that need them. From S1 onward, incoming skills similar to but distinct from earlier required skills also join. \ours{} refines these incoming skills, and the required skills keep their original metadata. After each session from S1 to S9, a checkpoint routes all queries released so far, so routing is tested at nine points as the library grows. The main study uses 198 BigCodeBench~\citep{zhuo2025bigcodebench} queries of SRA-Bench, each needing three to five skills, and its library grows from 27 to 731 skills (Appendix~\ref{app:benchmark} and Tab.~\ref{tab:sessions}). For RQ3, SkillRet and SkillRouter add 60 and 42 queries. RQ3 also installs the libraries in Codex CLI at every session (Appendix~\ref{app:rq3_supp}).

\paragraph{Router.}
As in agent harnesses such as Codex CLI~\citep{openai2025codexcli}, an LLM router selects skills from their names and descriptions alone. The main router is Claude Haiku 4.5~\citep{anthropic2025haiku45}, and it routes each query in two calls. The first call shortlists ten skills from all installed skills, and the second reorders the shortlist. The first five skills of the reordered list form the final selection, which is $\mathrm{top}_K$ in Eq.~\eqref{eq:reward} with $K=5$. We call this two-call pipeline our routing harness.

\paragraph{Metrics.}
To measure whether required skills remain selectable as the library grows, we report two metrics on the final selection $\mathrm{top}_K$. Full coverage at $K$ (FC@$K$), adopted from SkillRouter~\citep{zheng2026skillrouter}, is the fraction of queries whose required skills all appear in $\mathrm{top}_K$. It serves as the primary metric, because an agent needs every required skill to complete a query. Recall@$K$ is the fraction of each query's required skills that appear in $\mathrm{top}_K$, averaged over queries, and it shows how many required skills are lost when full coverage fails. We set $K=5$, because each query of the main study needs at most five skills and an agent loads only a few skills into its context. FC@5 at a checkpoint thus estimates $A_t$ in Eq.~\eqref{eq:reward}, and its average over the nine checkpoints S1--S9, reported as Avg, estimates the objective in Eq.~\eqref{eq:objective}.

\paragraph{Baselines.}
Besides \textbf{Vanilla}, which keeps the original metadata, we compare three kinds of baselines (Appendix~\ref{app:baselines}). The first kind rewrites the incoming skill. \textbf{Body-distilled}, adapted from SkillRouter~\citep{zheng2026skillrouter}, writes the incoming skill's description from that skill's body alone. The second kind rewrites earlier skills. \textbf{Neighbor-contrastive}, adapted from A-MEM's memory evolution~\citep{xu2025amem}, rewrites the three most similar earlier skills whenever a skill arrives. Like \ours{}, these rewriting baselines use the same generator and never see the evaluation queries or their labels. The third kind keeps the metadata and narrows the candidates. \textbf{Retrieve-and-rerank} (R\&R), SkillRouter's base pipeline, narrows the library to a short candidate list with an embedding retriever and a reranker~\citep{zhang2025qwen3embedding}, and the router's first call reads only these candidates. Its \emph{Hybrid} variants add BM25~\citep{robertson2009bm25} following \citet{liu2026skillswild}, and its \emph{-CFT} (continual fine-tuning) variants fine-tune the embedding model at every checkpoint on synthetic queries for the incoming skills~\citep{zheng2026skillrouter,murtaza2026synthetic}. We also run the released pages of Cluster-Contrastive Capability Pages (CC-Pages)~\citep{wang2026capability}.

\begin{figure*}[t]
\centering
\begin{minipage}[b]{0.32\textwidth}\centering
\includegraphics[width=\linewidth]{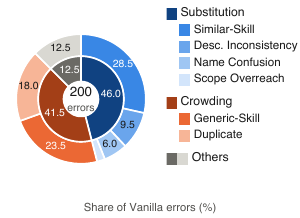}\\[1pt]
{\small (a) Vanilla errors by form and type}
\end{minipage}\hfill
\begin{minipage}[b]{0.32\textwidth}\centering
\includegraphics[width=\linewidth]{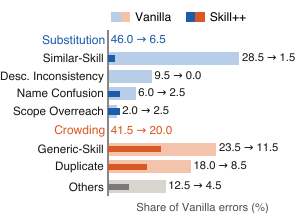}\\[1pt]
{\small (b) \ours{} against Vanilla}
\end{minipage}\hfill
\begin{minipage}[b]{0.32\textwidth}\centering
\includegraphics[width=\linewidth]{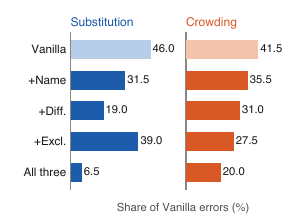}\\[1pt]
{\small (c) Components added to Vanilla}
\end{minipage}
\caption{This figure breaks down the routing errors at S9 into two forms and seven types (defined in Appendix~\ref{app:error_supp}), to show which errors \ours{} removes. All panels draw on the eight combinations of Tab.~\ref{tab:rq4-factorial}. (a)~The 200 Vanilla errors, by form (inner ring) and type (outer ring). (b)~Errors per type under Vanilla and \ours{}, as a share of the Vanilla errors. (c)~Errors per form when one component or all three are added to Vanilla, shown in the same way. \ours{} removes most errors of both forms, and each component mainly reduces one form.}
\label{fig:rq4-error}
\end{figure*}

\subsection{Experimental Results}
\label{sec:results}

\takeawaybox{RQ1: As the library grows, does \ours{} keep earlier skills selectable while new skills remain selected for their own queries?}{Takeaway 1:}{Yes. Refining only the incoming metadata against similar earlier skills prevents most of the loss caused by library growth and outperforms all baselines, including those that keep fine-tuning the retriever. New skills are selected for their own queries more often than under the original metadata, so \ours{} gains stability without trading away plasticity.}

Fig.~\ref{fig:routers}a shows that \textbf{the gap between \ours{} and the original metadata widens as the library grows}. From S1 to S9, FC@5 under Vanilla falls from 92.5\% to 36.4\%, while \ours{} still reaches 76.3\% and leads at every checkpoint. The strongest baseline reaches 67.8\%, and CC-Pages 60.1\%. As for plasticity, refined incoming skills rank first on 56.9\% of queries written from their bodies, compared with 51.9\% under Vanilla (Appendix~\ref{app:rq1_supp}).

\takeawaybox{RQ2: Which routing errors does \ours{} reduce, and does each component address the error it targets?}{Takeaway 2:}{Under the original metadata, most errors arise because a similar incoming skill takes the place of a required skill (substitution) or because generic and redundant skills fill the selected slots (crowding). \ours{} nearly eliminates substitution and halves crowding. Differentiation and name clarification mainly reduce substitution, and exclusions mainly reduce crowding.}

Fig.~\ref{fig:rq4-error} shows that \textbf{most Vanilla errors take one of two forms, and \ours{} reduces both}. Each required skill missing from the final selection at S9 counts as one error (Appendix~\ref{app:error_supp}).
\begin{expitems}
\item \textbf{Three types cause most Vanilla errors.} Substitution and crowding account for 46.0\% and 41.5\% of the 200 Vanilla errors (Fig.~\ref{fig:rq4-error}a). Similar-Skill Confusion, Generic-Skill Crowding and Duplicate Crowding together make up 70.0\%.
\item \textbf{\ours{} nearly removes substitution and halves crowding.} Substitution falls from 46.0\% to 6.5\% of the Vanilla errors and crowding from 41.5\% to 20.0\% (Fig.~\ref{fig:rq4-error}b). Similar-Skill Confusion falls from 28.5\% to 1.5\%, and Description Inconsistency, in which an incoming description overclaims or underclaims its body, disappears.
\item \textbf{Each component mainly reduces the form it targets.} Added alone to Vanilla, differentiation and name clarification lower substitution by 27.0 and 14.5~pp, while exclusions lower crowding by 14.0~pp (Fig.~\ref{fig:rq4-error}c). Each component lowers the other form by at most half as much.
\end{expitems}

\takeawaybox{RQ3: Does \ours{} generalize beyond SRA-Bench and Claude Haiku 4.5?}{Takeaway 3:}{Yes. Its gains generalize beyond the main study without retuning, because \ours{} changes only the metadata that every router reads. They persist on the SkillRet and SkillRouter benchmarks, across the \mbox{GPT-5.4-mini}, \mbox{GPT-5.6 Luna}, \mbox{GPT-5.6 Sol} and \mbox{Qwen3.7-Flash} routers, and in the Codex CLI harness. The gains are largest where routing over the original metadata is weakest, suggesting that refinement is most beneficial when a router alone cannot distinguish similar skills.}

Fig.~\ref{fig:routers}b--e and Fig.~\ref{fig:harness_nine} show that \textbf{\ours{} raises FC@5 in every new setting}. On SkillRet and SkillRouter (Tab.~\ref{tab:rq3_nine}), it stays above Vanilla at every session as the libraries grow to about 2,000 skills. Its mean FC@5 over S1--S9 is 10.9 and 12.4~pp higher, and it leads by 14.4 and 18.9~pp at S9.
Across the five routers, it stays above Vanilla at every checkpoint and improves S9 FC@5 by 23.1~pp on average (Fig.~\ref{fig:routers} and Appendix~\ref{app:rq3_supp}). At S9, it also exceeds Body-distilled across the four evaluated routers, from 31.9~pp with Claude Haiku 4.5 to 9.6~pp with GPT-5.6 Sol~\citep{openai2026gpt56}. In Codex CLI, its mean FC@5 over S1--S9 exceeds Vanilla by 34.3~pp with GPT-5.6 Luna and by 11.1~pp with GPT-5.6 Sol.

\takeawaybox{RQ4: How much does each component contribute?}{Takeaway 4:}{Every component contributes, and differentiation contributes most.}

Tab.~\ref{tab:rq2_factorial} shows that \textbf{removing any component lowers FC@5, and removing differentiation lowers it most}. At S9, FC@5 falls from 76.3\% to 56.6\%, 64.6\% and 61.6\% without differentiation, exclusions and name clarification, respectively.

%% file: sections/05_conclusion.tex
\section{Conclusion}
\label{sec:conclusion}

This paper formulates skill routing over a continually growing library as lifelong skill routing. New skills cause routing interference through substitution and crowding, even though neither the earlier skills nor the router changes. To mitigate this interference, we propose \ours{}, which refines the metadata of each incoming skill against similar earlier skills, without router training or labeled feedback. Experiments show that \ours{} outperforms all compared approaches and keeps both earlier and new skills selectable. It also nearly eliminates substitution and halves crowding, and its gains hold across benchmarks, routers and the Codex CLI harness. Future work can extend lifelong skill routing to multimodal skills and queries.

%% file: sections/A_implementation.tex
\section{Implementation Details}
\label{app:impl}

\begin{figure*}[!t]
  \centering
  \includegraphics[width=0.9\textwidth]{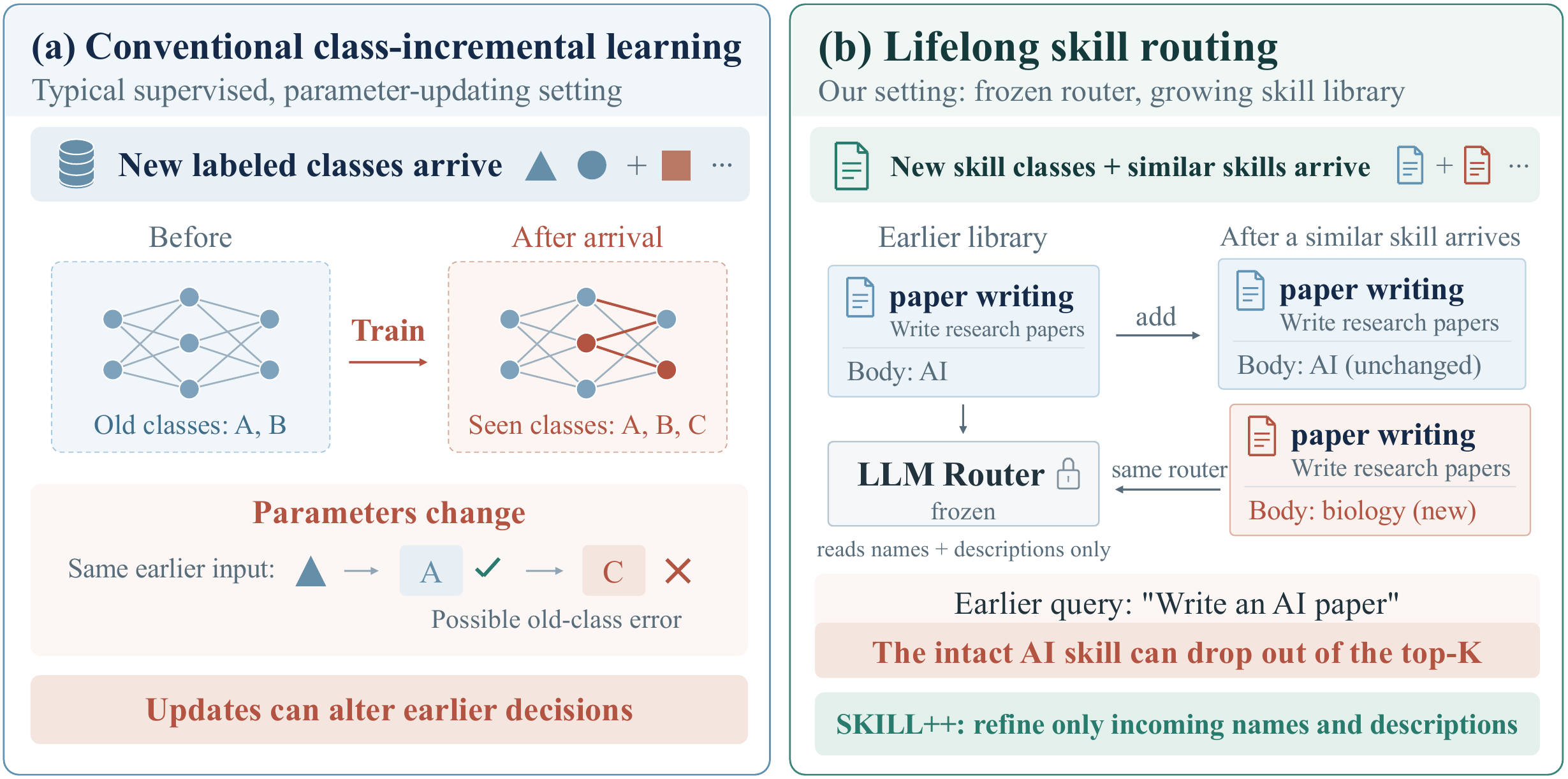}
  \caption{This figure compares conventional class-incremental learning with lifelong skill routing, to motivate our formulation. Both settings aim to retain earlier performance while accommodating new classes or skills. (a)~The model is trained on new labeled classes, so parameter updates can change its decisions on earlier classes. (b)~New skill classes and skills similar to earlier ones arrive while the router stays frozen and reads only names and descriptions. A biology paper-writing skill with the same metadata as the earlier AI skill can push the unchanged AI skill out of the top-$K$ for an earlier query, so \ours{} refines only the metadata of incoming skills.}
  \label{fig:cil}
\end{figure*}

This appendix gives the settings that Sec.~\ref{sec:method} defers. Fig.~\ref{fig:cil} first contrasts the formulation of Sec.~\ref{sec:formulation} with conventional class-incremental learning.

\paragraph{Description differentiation.}
The reference $g_t$ is selected with all-MiniLM-L6-v2~\citep{reimers2019sbert,wang2020minilm} embeddings of the original name and description. Gemini 3.5 Flash-Lite~\citep{google2026gemini35flashlite} then reads up to 1,600 characters of the incoming body and 1,200 of the reference body. Its prompt asks for at most 220 characters that state only capabilities contained in the incoming body, lead with what the reference does not offer, and never name or allude to the reference. Outputs that are too long or name the reference are regenerated and then retried with Gemini 3.6 Flash~\citep{google2026gemini36flash}, and the original description is kept if none passes.

\paragraph{Scope exclusions.}
For each skill, DeepSeek-V4-Flash~\citep{deepseek2026v4} writes up to eight synthetic queries from up to 2,600 characters of its body, phrased as a user would write them during a task and never naming the skill. Queries are generated once per skill and are checked for overlap with the 198 routing queries. Let $u(r)$ be the name and the first 300 characters of the description joined by a period, and let $\iota(q)$ prepend a retrieval instruction to a query. For a candidate $s$ with query set $\tilde Q_s$,
\begin{align}
\sigma_1(s)&=\max_{q\in \tilde Q_s}\psi(\iota(q))^{\top}\psi(u(r_t)),\label{eq:sigma1}\\
\sigma_2(s)&=\max_{q\in \tilde Q_s,\,q'\in \tilde Q_t}\psi(\iota(q))^{\top}\psi(\iota(q')),\label{eq:sigma2}\\
\sigma_3(s)&=\phi'(u(r_s))^{\top}\phi'(u(r_t)),\label{eq:sigma3}
\end{align}
where $\psi$ is Qwen3-Embedding-4B~\citep{zhang2025qwen3embedding} and $\phi'$ is MiniLM applied to $u(\cdot)$. Intuitively, $\sigma_1$ asks whether the incoming metadata matches a typical query of $s$, $\sigma_2$ whether the two skills serve similar queries, and $\sigma_3$ whether their names and descriptions look alike. The first two take a maximum because one shared use already suffices for confusion. Let $\rho_k(s)$ denote the rank of candidate $s$ under $\sigma_k$, with rank 1 indicating the highest similarity. Candidates are sorted lexicographically by
\begin{equation}
  \Bigl(\min_{k\in\{1,2,3\}}\rho_k(s),\;\sum_{k=1}^{3}\rho_k(s)\Bigr),
  \label{eq:fusion}
\end{equation}
with remaining ties resolved by addition order. This rule favors candidates that any single measure ranks highly and breaks ties by the sum of ranks. The first three candidates form $\mathcal{G}_t$. Gemini 3.5 Flash-Lite then writes one short lowercase phrase per reference from its name and the first 180 characters of its description. Phrases containing a reference's name head are dropped, and phrases are removed from the end until the revised description fits $\ell_{\max}=298$ characters.

\paragraph{Name clarification.}
Let $h(n)$ denote the name head of $n$, which is the lowercased first segment of $n$ after splitting on spaces, hyphens, or em dashes. We set $h(n)=\bot$ when this segment has at most three characters. Name heads come from the original names of earlier skills, and $z_t$ in Eq.~\eqref{eq:rename} flags a conflict. Containment is false when $h(n_s)=\bot$. Gemini 3.5 Flash-Lite reads up to 1,400 characters of the body together with $n_t$ and $n_{g_t}$, and proposes a name that adds a qualifier, such as a subtopic or type of guidance. The call is repeated up to three times until it returns a parsable proposal, which is used only if it has 4 to 60 characters. Otherwise the original name is kept.

\paragraph{Overall procedure.}
Algorithm~\ref{alg:skillpp} combines the three components into the procedure that \ours{} runs each time a skill is added (Sec.~\ref{sec:refine}). In the algorithm, $P$ lists the exclusion phrases in the order of $\mathcal{G}_t$, after phrases that contain a reference's name head are dropped. The clause $e(P)$ joins them after \texttt{Does not cover} and is empty when $P$ is empty. The symbol $\ell_{\max}$ denotes the length threshold for appending exclusion phrases. The algorithm first sets every output to its default (line~1). If earlier skills exist, it selects the reference skill $g_t$ and writes the capability description (lines 3--4). It then ranks related skills and writes one exclusion phrase for each (lines 5--6). The incoming skill is renamed only when its name contains the head of an earlier name (lines 7--8). Exclusion phrases are dropped from the end until the metadata fits $\ell_{\max}$ (lines 11--13). Finally, the algorithm assembles $\tilde r_t$ and stores the revised skill in the library and the original in the records (lines 14--16).

\begin{algorithm}[t]
\caption{\ours{} at addition step $t$}
\label{alg:skillpp}
\small
\begin{algorithmic}[1]
\Require incoming skill $c_t=((n_t,d_t),b_t)$; original-skill records $\mathcal{H}_{t-1}$; skill library $\mathcal{L}_{t-1}$; length threshold $\ell_{\max}$
\Ensure updated library $\mathcal{L}_t$; updated records $\mathcal{H}_t$
\State $\hat d_t\gets d_t$;\ \ $P\gets[\,]$;\ \ $\tilde n_t\gets n_t$ \Comment{defaults}
\If{$\mathcal{H}_{t-1}\neq\varnothing$}
  \State $g_t\gets$ most similar earlier skill by Eq.~\eqref{eq:reference}
  \Statex \hspace{\algorithmicindent}\hspace{\algorithmicindent}\Comment{reference skill for description and name}
  \State $\hat d_t\gets\mathcal{M}_D(n_t,b_t,n_{g_t},b_{g_t})$
  \Statex \hspace{\algorithmicindent}\hspace{\algorithmicindent}\Comment{capability description, $d_t$ if checks fail}
  \State $\mathcal{G}_t\gets$ up to three earlier skills ranked by Eq.~\eqref{eq:fusion}
  \Statex \hspace{\algorithmicindent}\hspace{\algorithmicindent}\Comment{reference set from synthetic queries}
  \State $P\gets\mathcal{M}_E\bigl(\{(n_s,d_s)\}_{s\in\mathcal{G}_t}\bigr)$
  \Statex \hspace{\algorithmicindent}\hspace{\algorithmicindent}\Comment{one exclusion phrase per reference skill}
  \If{$z_t=1$ by Eq.~\eqref{eq:rename}} \Comment{shared name head}
    \State $\tilde n_t\gets\mathcal{M}_N(n_t,b_t,n_{g_t})$ \Comment{$n_t$ if checks fail}
  \EndIf
\EndIf
\While{$|\hat d_t\oplus e(P)|>\ell_{\max}$ and $|P|>0$}
  \State remove the last phrase from $P$ \Comment{length threshold}
\EndWhile
\State $e_t\gets e(P)$;\ \ $\tilde r_t\gets(\tilde n_t,\ \hat d_t\oplus e_t)$ \Comment{Eq.~\eqref{eq:assemble}}
\State $\mathcal{L}_t\gets\mathcal{L}_{t-1}\cup\{(\tilde r_t,b_t)\}$ \Comment{Eq.~\eqref{eq:library-update}}
\State $\mathcal{H}_t\gets\mathcal{H}_{t-1}\cup\{c_t\}$ \Comment{original kept for later steps}
\State \Return $\mathcal{L}_t,\ \mathcal{H}_t$
\end{algorithmic}
\end{algorithm}

%% file: sections/B_experiments_supplement.tex
\section{Experimental Setup}
\label{app:exp_setup}

This appendix details the benchmark, the routing protocol and the baselines of Sec.~\ref{sec:setup}.

\begin{table*}[!t]
\centering
\small
\tbfit{\textwidth}{11}{l*{9}{N{00.0}}G{00.0}}{%
\toprule
Approach & \tbhead{S1} & \tbhead{S2} & \tbhead{S3} & \tbhead{S4} & \tbhead{S5} & \tbhead{S6} & \tbhead{S7} & \tbhead{S8} & \tbhead{S9} & \tbgaphead{Avg} \\
\midrule
\tbband{11}{FC@5 (\%)\,$\uparrow$: all required skills in the final selection} \\
Vanilla & \tbsecond{92.5} & 78.3 & 73.8 & 68.0 & 51.7 & 49.3 & 44.4 & 31.8 & 36.4 & 58.5 \\
Body-distilled & 90.0 & 80.0 & \tbsecond{86.2} & \tbsecond{72.0} & 68.3 & 57.1 & 55.6 & 47.5 & 44.4 & 66.8 \\
Neighbor-contrastive & \tbsecond{92.5} & \tbsecond{85.0} & 78.8 & 71.0 & 61.7 & 58.6 & 58.1 & 53.6 & 50.0 & 67.7 \\
R\&R & 90.0 & 80.0 & 75.0 & 70.0 & 62.5 & 60.7 & 60.0 & 62.0 & 58.6 & 68.8 \\
R\&R-CFT & 89.2 & 77.2 & 79.6 & 70.3 & \tbsecond{70.8} & 67.6 & 69.4 & 68.2 & \tbsecond{67.8} & 73.3 \\
Hybrid R\&R & \tbsecond{92.5} & 78.3 & 75.0 & 65.0 & 61.7 & 62.1 & 63.7 & 61.5 & 58.6 & 68.7 \\
Hybrid R\&R-CFT & 91.7 & 82.2 & 80.8 & 69.3 & 69.2 & \tbsecond{69.5} & \tbsecond{71.0} & \tbsecond{68.7} & 65.0 & \tbsecond{74.2} \\
CC-Pages & 85.0 & 68.3 & 72.5 & 66.0 & 65.8 & 64.3 & 58.1 & 62.0 & 60.1 & 66.9 \\
\tbours{\ours{} (ours)} & \tbbest{97.5} & \tbbest{98.3} & \tbbest{95.0} & \tbbest{85.0} & \tbbest{85.0} & \tbbest{79.3} & \tbbest{81.9} & \tbbest{78.8} & \tbbest{76.3} & \tbbest{86.3} \\
\midrule
\tbband{11}{Recall@5 (\%)\,$\uparrow$: share of required skills in the final selection} \\
Vanilla & 97.1 & 92.5 & 91.6 & 87.9 & 81.5 & 78.7 & 75.6 & 69.8 & 69.7 & 82.7 \\
Body-distilled & 96.5 & 93.5 & \tbsecond{95.5} & 89.5 & 86.8 & 83.0 & 82.1 & 79.3 & 75.6 & 86.9 \\
Neighbor-contrastive & \tbsecond{97.7} & \tbsecond{94.9} & 93.0 & 89.3 & 85.7 & 82.0 & 83.3 & 80.0 & 79.1 & 87.2 \\
R\&R & 96.5 & 93.2 & 91.5 & 89.6 & 86.1 & 85.4 & 85.1 & 85.4 & 83.5 & 88.5 \\
R\&R-CFT & 96.4 & 92.5 & 93.0 & \tbsecond{89.7} & \tbsecond{89.9} & 88.2 & 88.9 & 88.4 & \tbsecond{87.6} & 90.5 \\
Hybrid R\&R & 97.1 & 92.8 & 91.1 & 86.2 & 85.3 & 85.4 & 86.5 & 85.8 & 84.7 & 88.3 \\
Hybrid R\&R-CFT & 97.5 & 94.0 & 93.8 & 89.4 & 88.6 & \tbsecond{88.9} & \tbsecond{89.6} & \tbsecond{88.7} & 86.9 & \tbsecond{90.8} \\
CC-Pages & 94.2 & 89.7 & 90.8 & 88.6 & 87.8 & 86.6 & 85.0 & 86.2 & 84.5 & 88.2 \\
\tbours{\ours{} (ours)} & \tbbest{99.4} & \tbbest{99.6} & \tbbest{98.6} & \tbbest{94.5} & \tbbest{94.9} & \tbbest{93.4} & \tbbest{94.3} & \tbbest{92.3} & \tbbest{92.0} & \tbbest{95.4} \\
\bottomrule}
\caption{FC@5 is the share of queries whose required skills all reach the final selection, and Recall@5 is the share of each query's required skills that reach it, averaged over queries. The table gives both for every approach at every checkpoint of the main study with Claude Haiku 4.5. Fig.~\ref{fig:routers}a plots the FC@5 block, and -CFT results average three training seeds. Bold marks the best value per column and block, and underline the second.}
\label{tab:rq1_traj}
\end{table*}

\subsection{Benchmark Construction}
\label{app:benchmark}

\begin{table}[t]
\centering
\small
\tbfit{\columnwidth}{4}{lN{000}@{\,/\,}lN{0,000}@{\,/\,}lN{0,000}@{\,/\,}l}{%
\toprule
\tbstub{Session} & \multicolumn{6}{c}{Skills in the library / evaluated queries} \\
\cmidrule(l){2-7}
 & \multicolumn{2}{c}{SRA-Bench} & \multicolumn{2}{c}{SkillRet} & \multicolumn{2}{c}{SkillRouter} \\
\midrule
S1 & 87 & 40 & 83 & 12 & 82 & 8 \\
S2 & 152 & 60 & 200 & 18 & 207 & 11 \\
S3 & 224 & 80 & 319 & 24 & 339 & 15 \\
S4 & 305 & 100 & 483 & 30 & 535 & 19 \\
S5 & 387 & 120 & 725 & 36 & 758 & 23 \\
S6 & 470 & 140 & 996 & 42 & 1,015 & 27 \\
S7 & 556 & 160 & 1,242 & 48 & 1,316 & 30 \\
S8 & 644 & 179 & 1,552 & 54 & 1,610 & 33 \\
S9 & 731 & 198 & 1,954 & 60 & 1,982 & 37 \\
\bottomrule}
\caption{This table gives the library size and the number of evaluated queries at every checkpoint, to compare how the three benchmarks grow. The evaluated queries are all queries released so far.}
\label{tab:sessions}
\end{table}

Incoming skills come from the 25,626 web-collected skills of SRA-Bench and must pass three checks.
\begin{itemize}
\item \textbf{Similar.} The name and description lie close to those of the required skill in the embedding space of all-MiniLM-L6-v2, with a cosine similarity between 0.35 and 0.95.
\item \textbf{Confusable.} Gemini 3.5 Flash-Lite reads only names and descriptions and ranks the candidate among the two most relevant of four skills for at least one of two queries written for the required skill.
\item \textbf{Not a substitute.} GPT-5.5~\citep{openai2026gpt55} reads a query and both bodies and never judges that the candidate can replace the required skill.
\end{itemize}
Near-duplicate candidates are then removed. We keep the BigCodeBench queries that need three or more skills and have at least 20 such candidates for every required skill, which gives 198 queries.

The 198 queries are shuffled once and released in ten sessions of 19 or 20 queries. Each session first installs the required skills of its new queries and then adds up to two candidates for every required skill installed earlier, so the library grows from 27 skills at S0 to 731 at S9 (Tab.~\ref{tab:sessions}), and 679 of them are incoming skills. Each skill class starts with one required skill, and every incoming skill joins the class of the required skill it resembles. For example, the guide to Python's \texttt{math} module is joined by \texttt{math-tools} and \texttt{math} at S1 and by \texttt{math-help} and \texttt{discover-math} at S2. The library also holds seven skills that only 15 diagnostic queries outside the 198 need.

Within a session, skills are added one at a time, so each addition is one step $t$ of Sec.~\ref{sec:formulation}. A session thus spans many steps, and each refinement also sees the skills added earlier in the same session.

\subsection{Routing Protocol and Metrics}
\label{app:protocol}

All routers run at medium effort. They read up to 1,500 characters of the query and the first 300 characters of each description, in a fixed display order per query and checkpoint, and retrieval pipelines show their 50 candidates in the same order. The first call returns a shortlist of ten skills, and the second reorders it into the final ranking, which we score against the required skills $S^\star(q)$ of each query $q$.
\begin{itemize}
\item \textbf{FC@$K$} is 1 if all of $S^\star(q)$ are in $\mathrm{top}_K$.
\item \textbf{Recall@$K$} is the fraction of $S^\star(q)$ in $\mathrm{top}_K$.
\end{itemize}
Metrics are averaged over queries, count annotated skills and give no credit to substitutes. Confidence intervals (CIs) come from a paired bootstrap~\citep{koehn2004significance} over queries with 10,000 resamples at the 95\% level unless stated otherwise. Differences and averages use unrounded values.

\subsection{Baselines}
\label{app:baselines}

This section gives the inputs, models and settings of each baseline. The rewriting baselines use the generator of \ours{}, Gemini 3.5 Flash-Lite, keep the original names, and see no evaluation queries or labels.
\begin{itemize}
\item \textbf{Body-distilled.} Rewrites the description of each incoming skill from its own name and body and sees no other skill.
\item \textbf{Neighbor-contrastive.} At each addition, finds the three earlier skills whose metadata is most similar to the incoming skill and rewrites their descriptions to contrast with the incoming skill, whose body the generator reads. The incoming skill keeps its original metadata. The rewrites change the descriptions of 42 of the 45 required skills.
\item \textbf{R\&R.} Keeps the original metadata and narrows the library before the first call of the router. Qwen3-Embedding-0.6B embeds the query and the original name and description of every skill and retrieves the top 100 skills. Qwen3-Reranker-0.6B~\citep{zhang2025qwen3embedding} reorders them, and the first call reads the top 50. Neither model reads skill bodies. Two kinds of variants change this pipeline, and Hybrid R\&R-CFT combines both.
\begin{itemize}
\item \textbf{Hybrid R\&R.} Also ranks the top 100 skills by BM25~\citep{robertson2009bm25} over names and descriptions. Weighted reciprocal rank fusion~\citep{cormack2009rrf} with $k=0$ gives weight 0.8 to the embedding ranking and 0.2 to BM25, and the reranker then reorders the fused top 100.
\item \textbf{R\&R-CFT.} Fine-tunes the embedding model at every checkpoint on synthetic queries for the newly added skills, drawn from a pool of 6,008 queries that excludes all evaluation queries. Training uses an InfoNCE loss~\citep{oord2018cpc} with hard negatives and a replay memory of 512 earlier queries, and each update runs one epoch with batch size 16 and learning rate $10^{-5}$. Results average three training seeds within each query. The reranker stays frozen, because fine-tuning it as well lowers FC@5.
\end{itemize}
\item \textbf{CC-Pages.} We use the pages that the authors of this method~\citep{wang2026capability} released for SRA-Bench, unchanged. Each page adds positive triggers, negative boundaries and a discriminative body to a skill, compiled against the other skills of its cluster. Following the released code, Qwen3-Embedding-8B retrieves the ten closest skills of the current library, and Claude Haiku 4.5 reorders them in one call from their original names and descriptions with the negative boundaries. The pages were compiled over the full corpus, so they contrast skills that a session has not yet added, which favors this baseline before S9.
\end{itemize}

\section{Evidence for the Research Questions}
\label{app:exp_supp}

This appendix gives the evidence for each RQ in the order of Sec.~\ref{sec:results} and ends with checks of the routing protocol.

\subsection{RQ1: Stability and Plasticity as the Library Grows}
\label{app:rq1_supp}

\paragraph{Full results.}
Tab.~\ref{tab:rq1_traj} lists FC@5 and Recall@5 of every approach at every checkpoint, and its FC@5 block is plotted in Fig.~\ref{fig:routers}a. At S9, \ours{} exceeds Vanilla by 39.9~pp and the strongest baseline, R\&R-CFT, by 8.4~pp, with 95\% CIs of 32.3 to 47.5 and 2.4 to 14.5. CC-Pages stays below \ours{} at every session and reaches 60.1\% at S9. Most of its loss occurs at retrieval, where only ten skills reach the router.

\begin{figure}[!t]
\centering
\includegraphics[width=\columnwidth]{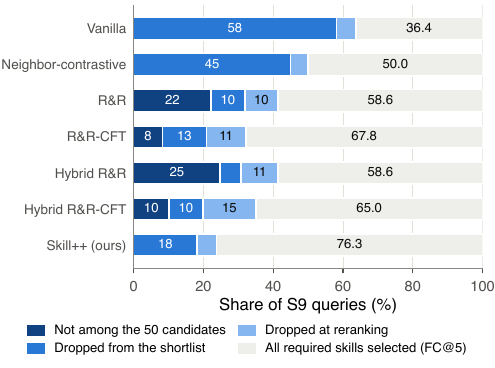}
\caption{Each bar splits the S9 queries of one approach by the stage at which they first lose a required skill. The skill is not among the 50 candidates from retrieval, which applies only to the retrieval baselines, or it is dropped from the shortlist or at reranking. The gray remainder, labeled with its value, holds the queries with all required skills selected and equals FC@5. Most failed queries lose a required skill before reranking.}
\label{fig:rq1_mechanism}
\end{figure}

\paragraph{Where required skills are lost.}
Fig.~\ref{fig:rq1_mechanism} shows that most losses under Vanilla occur when the router selects its shortlist. At S9, 58.1\% of queries lose a required skill at this stage and 5.6\% more at the rerank. The loss affects queries that previously succeeded, since 62 of the 115 queries routed correctly when their required skills first became available fail by S9, against 33 of 177 under \ours{}. Retrieval baselines only change which candidates the router reads. Fine-tuning raises the share of queries whose 50 candidates contain every required skill from 77.8\% to 91.8\%, yet the router, reading the same original metadata, still misses a required skill for about a quarter of them.

\begin{table*}[!t]
\centering
\small
\setlength{\tbgap}{0.3em}
\tbfit{\textwidth}{15}{cccD{00.0}{\,+00.0}D{00.0}{\,$-$00.0}*{5}{N{00.0}}D{00.0}{\,$-$00.0}N{00.0}N{00.0}D{00.0}{\,$-$00.0}N{100.0}}{%
\toprule
 & & & & \multicolumn{11}{c}{Missing required skills (\% of the 200 Vanilla errors)\,$\downarrow$} \\
\cmidrule(l){5-15}
\multicolumn{3}{c}{Components} & & \multicolumn{6}{c}{Substitution} & \multicolumn{3}{c}{Crowding} & & \\
\cmidrule(lr){1-3}\cmidrule(lr){5-10}\cmidrule(lr){11-13}
Diff. & Excl. & Name & \tbgaphead{FC@5\,$\uparrow$} & \tbgaphead{All} & \tbhead{Similar} & \tbhead{\shortstack{Same\\name}} & \tbhead{Desc.} & \tbhead{Scope} & \tbhead{Name} & \tbgaphead{All} & \tbhead{Duplicate} & \tbhead{Generic} & \tbgaphead{Others} & \tbhead{Total} \\
\midrule
& &  & 36.4\tbd{} & 46.0\tbd{} & 28.5 & 19.0 & 9.5 & \tbbest{2.0} & 6.0 & 41.5\tbd{} & 18.0 & 23.5 & 12.5\tbd{} & 100.0 \\
\addlinespace[2pt]
\checkmark & &  & 53.5\tbd{+17.2} & 19.0\tbd{$-$27.0} & 8.0 & 7.0 & 2.5 & 5.0 & 3.5 & 31.0\tbd{$-$10.5} & 15.0 & 16.0 & 9.5\tbd{$-$3.0} & 59.5 \\
& \checkmark & & 49.5\tbd{+13.1} & 39.0\tbd{$-$7.0} & 19.5 & 15.0 & 8.0 & 3.0 & 8.5 & 27.5\tbd{$-$14.0} & \tbsecond{10.0} & 17.5 & 10.5\tbd{$-$2.0} & 77.0 \\
& & \checkmark & 42.9\tbd{+6.6} & 31.5\tbd{$-$14.5} & 12.0 & 6.0 & 8.0 & 3.5 & 8.0 & 35.5\tbd{$-$6.0} & 16.5 & 19.0 & 10.0\tbd{$-$2.5} & 77.0 \\
\addlinespace[2pt]
\checkmark & \checkmark & & 61.6\tbd{+25.3} & 14.5\tbd{$-$31.5} & \tbsecond{5.0} & 5.0 & \tbsecond{1.0} & 4.0 & 4.5 & 28.5\tbd{$-$13.0} & 12.0 & 16.5 & \tbsecond{5.5}\tbd{$-$7.0} & 48.5 \\
\checkmark & & \checkmark & \tbsecond{64.6}\tbd{+28.3} & \tbsecond{13.5}\tbd{$-$32.5} & 5.5 & \tbsecond{3.5} & 1.5 & 3.5 & \tbsecond{3.0} & \tbsecond{23.0}\tbd{$-$18.5} & 13.0 & \tbbest{10.0} & 7.0\tbd{$-$5.5} & \tbsecond{43.5} \\
& \checkmark & \checkmark & 56.6\tbd{+20.2} & 17.5\tbd{$-$28.5} & 9.0 & 6.0 & 3.0 & 3.0 & \tbbest{2.5} & 30.0\tbd{$-$11.5} & 12.0 & 18.0 & 6.0\tbd{$-$6.5} & 53.5 \\
\addlinespace[2pt]
\rowcolor{tbours}\checkmark & \checkmark & \checkmark & \tbbest{76.3}\tbd{+39.9} & \tbbest{6.5}\tbd{$-$39.5} & \tbbest{1.5} & \tbbest{1.0} & \tbbest{0.0} & \tbsecond{2.5} & \tbbest{2.5} & \tbbest{20.0}\tbd{$-$21.5} & \tbbest{8.5} & \tbsecond{11.5} & \tbbest{4.5}\tbd{$-$8.0} & \tbbest{31.0} \\
\bottomrule}
\caption{This table routes all eight combinations of the three components at S9 with Claude Haiku 4.5, to show how they interact. A check mark turns a component on, so the first row is Vanilla and the shaded row is \ours{}. The error columns count the missing required skills by the forms and types of Fig.~\ref{fig:rq4-error}, as a share of the 200 Vanilla errors. Gray numbers give the change from Vanilla in points. Bold marks the best value per column and underline the second. Diff.: description differentiation. Excl.: scope exclusions. Name: name clarification, or Name Confusion among the error types. Desc.: Description Inconsistency. Same name: Similar-Skill Confusion errors in which the incoming skill has the name of the required skill. All configurations reuse the frozen outputs of \ours{}, and the exclusions-only row appends its clause to the original description.}
\label{tab:rq4-factorial}
\label{tab:rq2_factorial}
\end{table*}

\paragraph{Incoming skills.}
To test plasticity, we route 1,864 queries from the bodies of 673 refined incoming skills over the 731-skill library. The incoming skill ranks first on 56.9\% of its queries with \ours{} and on 51.9\% with Vanilla (+5.1~pp, 97.5\% CI 2.1 to 8.1).

\subsection{RQ2: Error Types and Counts}
\label{app:error_supp}

\paragraph{Error types.}
Each required skill missing from the final selection at S9 receives one type. Possible substitutes go to Others first, and the remaining errors receive the first matching type below. The types fall into two forms. In \emph{substitution}, a selected incoming skill takes the place of the missing skill. In \emph{crowding}, no selected skill resembles the missing skill, but other skills fill the slots.
\begin{itemize}
\item \textbf{Similar-Skill Confusion} (substitution). An incoming skill for the same work, usually with the same name, takes its place, for example an incoming \texttt{matplotlib} skill instead of the matplotlib guide that the query needs.
\item \textbf{Scope Overreach} (substitution). A broader skill, such as a toolkit or a general guide, takes the place of a dedicated skill, for example \texttt{data-viz-plots} while matplotlib and seaborn are left out.
\item \textbf{Name Confusion} (substitution). An incoming skill that matches only by name takes its place, for example a \texttt{math} skill that connects to a math tool instead of Python's math module.
\item \textbf{Description Inconsistency} (substitution). An incoming skill that resembles the missing skill by function takes its place, and its description misstates its own body. The description may overclaim, advertising work or a scope that the body does not support, or underclaim, omitting the main purpose of the body. For example, an incoming \texttt{matplotlib} skill promises animation that its body never covers. This type is checked before Similar-Skill Confusion. GPT-5.4-mini~\citep{openai2026gpt54mini} and DeepSeek-V4-Flash read the full body, and GPT-5.5 settles disagreements.
\item \textbf{Duplicate Crowding} (crowding). Two selected skills cover the same other part of the query and push the required skill out. This includes a broader skill selected next to the skill responsible for the same work.
\item \textbf{Generic-Skill Crowding} (crowding). A general-purpose skill that the query does not need, such as \texttt{python-best-practices}, takes a slot.
\item \textbf{Others.} A possible substitute, an incoming skill that some judges consider a substitute, takes its place, or no type above applies.
\end{itemize}

\begin{table*}[!t]
\centering
\small
\tbfit{\textwidth}{11}{l*{3}{N{100.0}}*{6}{N{00.0}}G{00.0}}{%
\toprule
\tbstub{Approach} & \multicolumn{10}{c}{FC@5 (\%)\,$\uparrow$} \\
\cmidrule(l){2-11}
 & \tbhead{S1} & \tbhead{S2} & \tbhead{S3} & \tbhead{S4} & \tbhead{S5} & \tbhead{S6} & \tbhead{S7} & \tbhead{S8} & \tbhead{S9} & \tbgaphead{Avg} \\
\midrule
\tbband{11}{Router GPT-5.4-mini} \\
Vanilla & 87.5 & 78.3 & 76.2 & 68.0 & \tbsecond{72.5} & \tbsecond{67.9} & 61.3 & 62.6 & 59.6 & 70.4 \\
Neighbor-contrastive & 85.0 & 75.0 & 75.0 & 62.0 & 62.5 & 57.9 & 56.9 & 54.7 & 57.6 & 65.2 \\
R\&R & 92.5 & 80.0 & 75.0 & 62.0 & 66.7 & 55.0 & 59.4 & 59.2 & 52.0 & 66.9 \\
R\&R-CFT & \tbsecond{94.2} & \tbsecond{82.8} & \tbsecond{82.9} & 70.0 & 69.2 & 66.0 & \tbsecond{67.3} & \tbsecond{66.9} & \tbsecond{65.2} & \tbsecond{73.8} \\
Hybrid R\&R & 87.5 & 78.3 & 72.5 & 67.0 & 63.3 & 56.4 & 56.9 & 59.2 & 53.0 & 66.0 \\
Hybrid R\&R-CFT & 89.2 & \tbsecond{82.8} & 82.1 & \tbsecond{71.0} & 67.8 & \tbsecond{67.9} & 65.8 & 66.7 & 62.5 & 72.8 \\
\tbours{\ours{} (ours)} & \tbbest{100.0} & \tbbest{100.0} & \tbbest{97.5} & \tbbest{92.0} & \tbbest{89.2} & \tbbest{87.1} & \tbbest{86.2} & \tbbest{85.5} & \tbbest{86.9} & \tbbest{91.6} \\
\midrule
\tbband{11}{Router GPT-5.6 Luna} \\
Vanilla & 90.0 & 85.0 & \tbsecond{86.2} & \tbsecond{82.0} & \tbsecond{77.5} & 72.9 & \tbsecond{75.0} & \tbsecond{73.2} & 71.7 & \tbsecond{79.3} \\
Neighbor-contrastive & 85.0 & 76.7 & 80.0 & 78.0 & 73.3 & 71.4 & 70.6 & 69.3 & \tbsecond{72.2} & 75.2 \\
R\&R & \tbsecond{92.5} & 85.0 & 77.5 & 76.0 & 70.0 & 64.3 & 60.6 & 62.6 & 62.6 & 72.3 \\
R\&R-CFT & 91.7 & \tbsecond{85.6} & 82.9 & 78.3 & 76.1 & \tbsecond{74.0} & 73.1 & 73.0 & 70.2 & 78.3 \\
Hybrid R\&R & \tbbest{95.0} & 83.3 & 75.0 & 74.0 & 66.7 & 60.0 & 65.0 & 59.8 & 58.6 & 70.8 \\
Hybrid R\&R-CFT & \tbsecond{92.5} & 83.9 & 84.2 & 75.7 & 75.8 & 71.7 & 72.9 & 72.4 & 69.4 & 77.6 \\
\tbours{\ours{} (ours)} & \tbbest{95.0} & \tbbest{96.7} & \tbbest{100.0} & \tbbest{91.0} & \tbbest{90.8} & \tbbest{86.4} & \tbbest{83.8} & \tbbest{84.9} & \tbbest{85.9} & \tbbest{90.5} \\
\midrule
\tbband{11}{Router GPT-5.6 Sol} \\
Vanilla & 92.5 & 88.3 & \tbsecond{90.0} & \tbsecond{85.0} & 85.0 & \tbsecond{83.6} & \tbsecond{81.9} & \tbsecond{84.4} & 76.8 & \tbsecond{85.3} \\
Neighbor-contrastive & 90.0 & 88.3 & 86.2 & 81.0 & \tbsecond{87.5} & 80.0 & \tbsecond{81.9} & 78.8 & \tbsecond{81.3} & 83.9 \\
R\&R & \tbbest{97.5} & 86.7 & 82.5 & 79.0 & 78.3 & 65.7 & 68.1 & 64.2 & 62.6 & 76.1 \\
R\&R-CFT & \tbsecond{94.2} & \tbsecond{90.0} & 88.3 & 81.3 & 81.4 & 75.2 & 74.6 & 76.5 & 73.7 & 81.7 \\
Hybrid R\&R & 92.5 & 85.0 & 81.2 & 78.0 & 76.7 & 65.7 & 66.2 & 66.5 & 60.1 & 74.7 \\
Hybrid R\&R-CFT & \tbsecond{94.2} & \tbsecond{90.0} & 88.8 & 78.7 & 81.4 & 73.1 & 74.6 & 75.4 & 73.4 & 81.1 \\
\tbours{\ours{} (ours)} & \tbbest{97.5} & \tbbest{98.3} & \tbbest{97.5} & \tbbest{89.0} & \tbbest{90.0} & \tbbest{90.0} & \tbbest{89.4} & \tbbest{88.3} & \tbbest{88.4} & \tbbest{92.0} \\
\midrule
\tbband{11}{Router Qwen3.7-Flash} \\
Vanilla & 85.0 & 78.3 & 75.0 & 62.0 & 59.2 & 55.0 & 59.4 & 56.4 & 49.0 & 64.4 \\
Neighbor-contrastive & 82.5 & \tbsecond{81.7} & \tbsecond{81.2} & 68.0 & 68.3 & 65.7 & 55.6 & 58.1 & 61.6 & 69.2 \\
R\&R & 80.0 & 80.0 & \tbsecond{81.2} & 66.0 & 65.8 & 62.9 & 63.7 & 63.7 & 56.6 & 68.9 \\
R\&R-CFT & 87.5 & 78.3 & 74.6 & \tbsecond{71.0} & \tbsecond{71.9} & \tbsecond{69.5} & \tbsecond{67.5} & 65.9 & \tbsecond{64.3} & \tbsecond{72.3} \\
Hybrid R\&R & \tbsecond{90.0} & 78.3 & 71.2 & 61.0 & 60.8 & 58.6 & 63.7 & 55.9 & 53.5 & 65.9 \\
Hybrid R\&R-CFT & 83.3 & 80.6 & 80.4 & 67.0 & 69.2 & 66.7 & \tbsecond{67.5} & \tbsecond{67.2} & 64.0 & 71.8 \\
\tbours{\ours{} (ours)} & \tbbest{97.5} & \tbbest{91.7} & \tbbest{96.2} & \tbbest{84.0} & \tbbest{85.0} & \tbbest{82.9} & \tbbest{76.9} & \tbbest{74.3} & \tbbest{71.7} & \tbbest{84.5} \\
\bottomrule}
\caption{This table reports FC@5 (\%) of the four other routers at every checkpoint and their mean over S1--S9 (Avg), to test whether the results of the main router hold. Each router routes the approaches of Tab.~\ref{tab:rq1_traj} except Body-distilled and CC-Pages, on the same inputs. -CFT results average three training seeds. Bold marks the best value per column within each router, and underline the second. At S9, \ours{} exceeds the strongest other approach in this table by 21.7, 13.6, 7.1 and 7.4~pp with GPT-5.4-mini, GPT-5.6 Luna, GPT-5.6 Sol and Qwen3.7-Flash.}
\label{tab:routers}
\end{table*}

\begin{table*}[!t]
\centering
\small
\tbfit{\textwidth}{11}{l*{3}{N{100.0}}*{6}{N{00.0}}G{00.0}}{%
\toprule
\tbstub{Approach} & \multicolumn{10}{c}{Recall@5 (\%)\,$\uparrow$} \\
\cmidrule(l){2-11}
 & \tbhead{S1} & \tbhead{S2} & \tbhead{S3} & \tbhead{S4} & \tbhead{S5} & \tbhead{S6} & \tbhead{S7} & \tbhead{S8} & \tbhead{S9} & \tbgaphead{Avg} \\
\midrule
\tbband{11}{Router GPT-5.4-mini} \\
Vanilla & 95.6 & 92.8 & 92.1 & 86.8 & \tbsecond{89.6} & 86.5 & 83.7 & 84.3 & 82.1 & 88.2 \\
Neighbor-contrastive & 94.2 & 91.1 & 90.4 & 82.9 & 84.3 & 81.4 & 81.0 & 79.5 & 80.6 & 85.0 \\
R\&R & 97.7 & 92.2 & 91.6 & 85.4 & 87.0 & 82.8 & 85.2 & 82.8 & 81.2 & 87.3 \\
R\&R-CFT & \tbsecond{98.3} & \tbsecond{94.3} & \tbsecond{94.1} & 88.8 & 88.4 & \tbsecond{86.9} & \tbsecond{88.0} & \tbsecond{87.6} & \tbsecond{86.5} & \tbsecond{90.3} \\
Hybrid R\&R & 95.4 & 93.2 & 90.0 & 85.4 & 86.4 & 83.2 & 83.3 & 83.4 & 81.7 & 86.9 \\
Hybrid R\&R-CFT & 96.5 & 94.0 & 93.4 & \tbsecond{89.3} & 87.8 & \tbsecond{86.9} & 87.6 & 87.2 & 85.9 & 89.8 \\
\tbours{\ours{} (ours)} & \tbbest{100.0} & \tbbest{100.0} & \tbbest{99.3} & \tbbest{97.8} & \tbbest{96.8} & \tbbest{95.5} & \tbbest{95.7} & \tbbest{95.2} & \tbbest{95.6} & \tbbest{97.3} \\
\midrule
\tbband{11}{Router GPT-5.6 Luna} \\
Vanilla & 96.2 & 94.7 & \tbsecond{95.9} & \tbsecond{93.1} & \tbsecond{92.1} & 89.6 & \tbsecond{90.0} & 89.6 & \tbsecond{89.6} & \tbsecond{92.3} \\
Neighbor-contrastive & 94.4 & 91.8 & 92.2 & 91.8 & 90.8 & 88.5 & 88.2 & 88.0 & 88.5 & 90.5 \\
R\&R & 96.5 & 93.1 & 92.5 & 89.8 & 89.2 & 86.9 & 84.6 & 85.3 & 85.5 & 89.3 \\
R\&R-CFT & 96.2 & \tbsecond{95.1} & 94.1 & 91.5 & 91.7 & \tbsecond{89.7} & 89.7 & \tbsecond{90.2} & 88.9 & 91.9 \\
Hybrid R\&R & \tbsecond{97.3} & 93.2 & 92.1 & 89.4 & 87.7 & 83.5 & 85.6 & 84.0 & 83.7 & 88.5 \\
Hybrid R\&R-CFT & 96.5 & 93.6 & 94.3 & 90.1 & 91.4 & 88.9 & 89.5 & 89.9 & 88.2 & 91.4 \\
\tbours{\ours{} (ours)} & \tbbest{98.5} & \tbbest{99.0} & \tbbest{100.0} & \tbbest{97.4} & \tbbest{96.4} & \tbbest{94.9} & \tbbest{94.4} & \tbbest{94.5} & \tbbest{95.2} & \tbbest{96.7} \\
\midrule
\tbband{11}{Router GPT-5.6 Sol} \\
Vanilla & 97.1 & 95.8 & \tbsecond{96.9} & \tbsecond{95.0} & \tbsecond{95.2} & \tbsecond{94.7} & \tbsecond{93.6} & \tbsecond{95.0} & 92.3 & \tbsecond{95.1} \\
Neighbor-contrastive & 94.8 & 95.3 & 95.8 & 93.1 & 94.7 & 92.0 & 93.5 & 92.0 & \tbsecond{93.6} & 93.9 \\
R\&R & \tbsecond{98.8} & 95.8 & 94.0 & 91.3 & 92.1 & 87.1 & 87.8 & 86.9 & 85.8 & 91.1 \\
R\&R-CFT & 97.8 & \tbsecond{96.5} & 95.9 & 92.5 & 93.6 & 90.8 & 90.2 & 91.7 & 90.5 & 93.3 \\
Hybrid R\&R & 97.1 & 94.4 & 93.4 & 90.7 & 90.1 & 86.7 & 87.5 & 87.3 & 85.2 & 90.3 \\
Hybrid R\&R-CFT & 98.1 & \tbsecond{96.5} & 96.2 & 92.0 & 93.4 & 90.0 & 90.8 & 91.0 & 90.4 & 93.2 \\
\tbours{\ours{} (ours)} & \tbbest{99.4} & \tbbest{99.6} & \tbbest{99.2} & \tbbest{96.6} & \tbbest{96.8} & \tbbest{96.6} & \tbbest{96.5} & \tbbest{96.1} & \tbbest{96.4} & \tbbest{97.5} \\
\midrule
\tbband{11}{Router Qwen3.7-Flash} \\
Vanilla & 95.4 & 93.5 & 91.3 & 86.5 & 85.4 & 83.6 & 83.6 & 82.2 & 79.2 & 86.7 \\
Neighbor-contrastive & 95.0 & \tbsecond{94.6} & \tbsecond{93.5} & 89.2 & 88.6 & 86.9 & 83.5 & 83.0 & 85.0 & 88.8 \\
R\&R & 93.5 & 93.1 & 93.3 & 88.2 & 87.8 & 85.3 & 86.1 & 86.1 & 83.3 & 88.5 \\
R\&R-CFT & \tbsecond{96.2} & 92.0 & 91.0 & \tbsecond{89.3} & \tbsecond{90.1} & \tbsecond{88.7} & 87.9 & 87.7 & \tbsecond{87.1} & \tbsecond{90.0} \\
Hybrid R\&R & 95.8 & 92.5 & 89.4 & 83.9 & 86.7 & 84.2 & 85.3 & 83.9 & 82.9 & 87.2 \\
Hybrid R\&R-CFT & 94.9 & 93.5 & 93.0 & 88.2 & 89.2 & 88.1 & \tbsecond{88.4} & \tbsecond{88.3} & 86.9 & \tbsecond{90.0} \\
\tbours{\ours{} (ours)} & \tbbest{99.4} & \tbbest{97.6} & \tbbest{98.5} & \tbbest{95.4} & \tbbest{95.1} & \tbbest{94.1} & \tbbest{91.9} & \tbbest{91.5} & \tbbest{90.7} & \tbbest{94.9} \\
\bottomrule}
\caption{This table reports Recall@5 (\%) of the four other routers at every checkpoint and its mean over S1--S9 (Avg), to show how many required skills each approach keeps with other routers. The routing runs are those of Tab.~\ref{tab:routers}, and -CFT results average three training seeds. Bold marks the best value per column within each router, and underline the second.}
\label{tab:routers_recall}
\end{table*}

\paragraph{Judging.}
Model judges assign the types in three steps. First, for every failed query, Gemini 3.5 Flash-Lite and DeepSeek-V4-Flash judge whether each selected incoming skill resembles a missing required skill. They read the query and both skills but not the routing result, and GPT-5.4-mini settles their disagreements. Second, the same judges decide from the skills alone whether a skill is broader or general-purpose. Third, a strict check asks whether a selected skill can in fact replace the missing one. GPT-5.6 Sol and Claude Opus 4.6~\citep{anthropic2026opus46} judge each such pair, and the skill counts as a verified substitute only if both consistently find that its body documents everything the query needs. A missing skill with a verified substitute is not counted as an error.

\paragraph{Counts.}
In Fig.~\ref{fig:rq4-error} and Tab.~\ref{tab:rq4-factorial}, \ours{} reduces the errors from 200 to 62.
\begin{itemize}
\item \textbf{Name clarification.} Name clarification alone reduces same-name Similar-Skill Confusion from 38 to 12 errors (19.0\% to 6.0\%). Changing only the names of the 81 colliding incoming skills to the clarified names raises the number of fully covered queries by 11, whereas a meaningless unique suffix lowers it by 1.
\item \textbf{Description Inconsistency.} Description Inconsistency accounts for 19 of the 200 Vanilla errors, 5 with differentiation alone and none under \ours{}. An audit of 85 incoming skills selected under Vanilla finds 18 descriptions that misstate the body, all by overclaiming.
\end{itemize}

\subsection{\texorpdfstring{RQ3: Other Benchmarks, Routers and Agent Harness}{RQ3: Other Benchmarks, Routers and Agent Harness}}
\label{app:rq3_supp}

\paragraph{Other benchmarks.}
We rebuild library growth on two public multi-skill benchmarks with the recipe, router and protocol of the main study.
\begin{itemize}
\item \textbf{Benchmarks.} \emph{SkillRet}~\citep{kang2026skillret} contributes 60 developer queries that each need three skills, and \emph{SkillRouter}~\citep{zheng2026skillrouter} contributes 42 multi-skill queries. FC@5 on SkillRouter covers the 37 queries with at most five required skills. Under three addition orders, each session adds as many incoming skills per required or diagnostic skill as the main study. Because SkillRet has 139 and SkillRouter 141 required skills, against 45 required and 7 diagnostic skills in the main study, their libraries grow to 1,954 and 1,982 at S9 on average (Tab.~\ref{tab:sessions}). The incoming skills pass the main study's checks.
\item \textbf{Results.} Vanilla falls from 88.9\% FC@5 at S1 to 15.6\% at S9 on SkillRet and from 80.1\% to 19.8\% on SkillRouter (Tab.~\ref{tab:rq3_nine}). \ours{} is higher at every session and ends at 30.0\% and 38.7\%, 14.4 and 18.9~pp above Vanilla.
\end{itemize}

\begin{table*}[!t]
\centering
\small
\tbfit{\textwidth}{12}{l*{9}{N{00.0}}G{00.0}G{+00.0}}{%
\toprule
\tbstub{Approach} & \multicolumn{10}{c}{FC@5 (\%)\,$\uparrow$} & \tbgaphead{\tbstub{$\Delta$ at S9}} \\
\cmidrule(lr){2-11}
 & \tbhead{S1} & \tbhead{S2} & \tbhead{S3} & \tbhead{S4} & \tbhead{S5} & \tbhead{S6} & \tbhead{S7} & \tbhead{S8} & \tbhead{S9} & \tbgaphead{Avg} & \\
\midrule
\tbband{12}{SkillRet} \\
Vanilla & 88.9 & 75.9 & 59.7 & 50.0 & 35.2 & 29.4 & 20.8 & 16.7 & 15.6 & 43.6 & \\
\tbours{\ours{} (ours)} & \tbbest{91.7} & \tbbest{90.7} & \tbbest{70.8} & \tbbest{65.6} & \tbbest{45.4} & \tbbest{38.9} & \tbbest{30.6} & \tbbest{27.2} & \tbbest{30.0} & \tbbest{54.5} & +14.4 \\
\midrule
\tbband{12}{SkillRouter} \\
Vanilla & 80.1 & 64.6 & 52.4 & 43.3 & 29.6 & 31.2 & 23.4 & 29.0 & 19.8 & 41.5 & \\
\tbours{\ours{} (ours)} & \tbbest{84.3} & \tbbest{67.9} & \tbbest{57.0} & \tbbest{50.3} & \tbbest{59.1} & \tbbest{43.8} & \tbbest{46.7} & \tbbest{37.0} & \tbbest{38.7} & \tbbest{53.9} & +18.9 \\
\bottomrule}
\caption{This table reports FC@5 (\%) at every session on SkillRet and SkillRouter, to test whether the gains hold on other benchmarks as their libraries grow to about 2,000 skills. Values are means over three addition orders, with Claude Haiku 4.5 as the router. Bold marks the better value per column. $\Delta$: \ours{} minus Vanilla.}
\label{tab:rq3_nine}
\end{table*}

\paragraph{Random skills.}
Three streams each add 679 skills sampled uniformly from the other skills of the SRA-Bench corpus to the 52 required and diagnostic skills. Vanilla then falls only from 97.0\% to 84.3\% FC@5, averaged over the streams. At S9, Body-distilled reaches 89.6\% and \ours{} 93.3\%, 8.9~pp above Vanilla.

\paragraph{Other routers.}
Fig.~\ref{fig:routers}b--e and Tab.~\ref{tab:routers} repeat the main comparison with GPT-5.4-mini, GPT-5.6 Luna, GPT-5.6 Sol and Qwen3.7-Flash~\citep{qwen2026qwen37flash} at every checkpoint, on the same inputs. With all five routers, \ours{} stays above Vanilla at every checkpoint (Tab.~\ref{tab:rq1_traj} and Tab.~\ref{tab:routers}). At S9 it gains 39.9, 22.7, 27.3, 14.1 and 11.6~pp with Claude Haiku 4.5, Qwen3.7-Flash, GPT-5.4-mini, GPT-5.6 Luna and GPT-5.6 Sol, ordered by Vanilla FC@5. This gain averages 23.1~pp, and all five 95\% paired CIs exclude zero. Body-distilled, routed at S9 with Claude Haiku 4.5, GPT-5.4-mini, GPT-5.6 Luna and GPT-5.6 Sol, trails \ours{} by 31.9, 23.8, 12.2 and 9.6~pp in the order just named, 19.4~pp on average. With all four other routers, \ours{} also has the highest Recall@5 at every checkpoint (Tab.~\ref{tab:routers_recall}). The approaches rank in the same order by average Recall@5 as by average FC@5, except that R\&R-CFT and Hybrid R\&R-CFT tie with Qwen3.7-Flash. The gaps are smaller in Recall@5, because a query that fails usually loses only some of its required skills. At S9, \ours{} leads Vanilla by 11.4, 13.5, 5.6 and 4.2~pp in Recall@5 with Qwen3.7-Flash, GPT-5.4-mini, GPT-5.6 Luna and GPT-5.6 Sol, against 22.7, 27.3, 14.1 and 11.6~pp in FC@5.

\begin{figure}[t]
\centering
\includegraphics[width=\columnwidth]{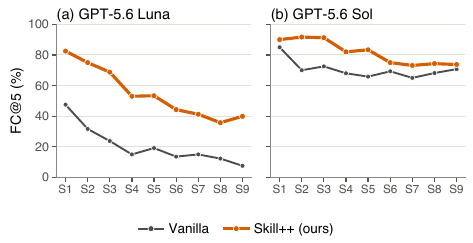}
\caption{This figure shows FC@5 at every session in Codex CLI with GPT-5.6 Luna and GPT-5.6 Sol, to test whether the gains reach an agent harness as the library grows (values in Tab.~\ref{tab:harness_nine}). \ours{} stays above Vanilla at every session with both models.}
\label{fig:harness_nine}
\end{figure}

\paragraph{Agent harness.}
We repeat the growth of the main study inside Codex CLI with GPT-5.6 Luna and GPT-5.6 Sol at medium effort.
\begin{itemize}
\item \textbf{Setup.} Each library is installed as native skill files, and the agent sees Codex CLI's own skill list of names and descriptions, its default tools and the original task prompt. We change no tools, prompts or routing logic and only raise the skill-list budget so that all 731 skills fit. At every session, each released query starts a fresh session, and we score the first five skills the agent names with FC@5.
\item \textbf{Selection.} With GPT-5.6 Luna, \ours{} leads Vanilla by 32.3~pp at S9 (95\% CI 25.8 to 38.9) and by 34.3~pp on average over S1--S9 (Tab.~\ref{tab:harness_nine} and Fig.~\ref{fig:harness_nine}). With GPT-5.6 Sol, which already selects well over the original metadata, \ours{} is higher at every session, by 11.1~pp on average and by 3.0~pp at S9.
\item \textbf{Task success.} We check task success on the 87-skill S1 library with 150 queries. There a workflow prompt asks the agent to name five skills before any tool call, and the hidden tests run after it writes its solution. \ours{} raises FC@5 from 33.3\% to 77.3\% with GPT-5.6 Luna and from 82.0\% to 88.0\% with GPT-5.6 Sol. Hidden-test pass rates also rise, by 4.0 and 0.7~pp.
\end{itemize}

\begin{table*}[!t]
\centering
\small
\tbfit{\textwidth}{12}{l*{9}{N{00.0}}G{00.0}G{+00.0}}{%
\toprule
\tbstub{Approach} & \multicolumn{10}{c}{FC@5 (\%)\,$\uparrow$} & \tbgaphead{\tbstub{$\Delta$ at S9}} \\
\cmidrule(lr){2-11}
 & \tbhead{S1} & \tbhead{S2} & \tbhead{S3} & \tbhead{S4} & \tbhead{S5} & \tbhead{S6} & \tbhead{S7} & \tbhead{S8} & \tbhead{S9} & \tbgaphead{Avg} & \\
\midrule
\tbband{12}{Codex CLI, GPT-5.6 Luna} \\
Vanilla & 47.5 & 31.7 & 23.8 & 15.0 & 19.2 & 13.6 & 15.0 & 12.3 & 7.6 & 20.6 & \\
\tbours{\ours{} (ours)} & \tbbest{82.5} & \tbbest{75.0} & \tbbest{68.8} & \tbbest{53.0} & \tbbest{53.3} & \tbbest{44.3} & \tbbest{41.2} & \tbbest{35.8} & \tbbest{39.9} & \tbbest{54.9} & +32.3 \\
\midrule
\tbband{12}{Codex CLI, GPT-5.6 Sol} \\
Vanilla & 85.0 & 70.0 & 72.5 & 68.0 & 65.8 & 69.3 & 65.0 & 68.2 & 70.7 & 70.5 & \\
\tbours{\ours{} (ours)} & \tbbest{90.0} & \tbbest{91.7} & \tbbest{91.2} & \tbbest{82.0} & \tbbest{83.3} & \tbbest{75.0} & \tbbest{73.1} & \tbbest{74.3} & \tbbest{73.7} & \tbbest{81.6} & +3.0 \\
\bottomrule}
\caption{This table reports FC@5 (\%) at every session in Codex CLI with GPT-5.6 Luna and GPT-5.6 Sol, to test whether the gains reach an agent harness as the library grows. FC@5 scores the first five skills that the agent names for the queries released so far, on the libraries of Tab.~\ref{tab:sessions}. Bold marks the better value per column. $\Delta$: \ours{} minus Vanilla.}
\label{tab:harness_nine}
\end{table*}

\subsection{RQ4: Ablation Study}
\label{app:ablation_supp}

\paragraph{All combinations.}
Tab.~\ref{tab:rq2_factorial} routes all eight combinations at S9, and adding a missing component raises FC@5 in all twelve cases, by 6.6 to 21.7~pp. Differentiation and exclusions together gain 25.3~pp over Vanilla, about 83\% of the sum of their separate gains of 17.2 and 13.1~pp, because differentiation mainly removes substitution and exclusions mainly remove crowding (Fig.~\ref{fig:rq4-error}c).

\paragraph{Reference choice.}
With all other inputs fixed, writing differentiation against the nearest earlier skill outperforms a random earlier skill by 9.5~pp (95\% CI 4.2 to 14.8) and an independent summary without references by 8.7~pp (95\% CI 3.0 to 14.4).

\subsection{Robustness}
\label{app:protocol_checks}

\begin{table}[!t]
\centering
\small
\tbfit{\columnwidth}{4}{lN{00.0}N{00.0}N{$-$0.0}}{%
\toprule
\tbstub{Protocol change} & \multicolumn{2}{c}{FC@5 (\%)\,$\uparrow$} & \tbhead{\tbstub{$\Delta$ (pp)}} \\
\cmidrule(lr){2-3}
 & \tbhead{Before} & \tbhead{After} & \\
\midrule
Display order shuffled & 85.9 & 86.9 & +1.0 \\
Identical inputs repeated & 85.9 & 84.8 & $-$1.0 \\
Addition order permuted & 82.3 & 79.3 & $-$3.0 \\
\bottomrule}
\caption{This table reruns \ours{} at S9 with GPT-5.6 Luna under three protocol changes and reports FC@5 before and after each change. $\Delta$: After minus Before. Every change shifts FC@5 by at most 3.0~pp, so routing noise stays small.}
\label{tab:protocol_checks}
\end{table}

To check that the gains are robust to the routing protocol, we rerun \ours{} at S9 with GPT-5.6 Luna under three changes to the routing protocol (Tab.~\ref{tab:protocol_checks}). Shuffling the display order tests whether the router favors skills listed early. Repeating identical inputs measures the randomness of the router itself. Permuting the addition order within each checkpoint changes the earlier skills that each refinement sees, so we rebuild \ours{} in both orders with GPT-5.6 Luna as the generator. The changes range from $-$3.0 to +1.0~pp. These shifts are small next to the 14.1~pp by which \ours{} outperforms Vanilla with the same router (Tab.~\ref{tab:routers}).

%% file: main.bbl
\begin{thebibliography}{57}
\providecommand{\natexlab}[1]{#1}

\bibitem[{{Anthropic}(2025)}]{anthropic2025haiku45}
{Anthropic}. 2025.
\newblock \href
  {https://www-cdn.anthropic.com/7aad69bf12627d42234e01ee7c36305dc2f6a970.pdf}
  {Claude {Haiku} 4.5 system card}.

\bibitem[{{Anthropic}(2026)}]{anthropic2026opus46}
{Anthropic}. 2026.
\newblock \href {https://anthropic.com/claude-opus-4-6-system-card} {Claude
  {Opus} 4.6 system card}.

\bibitem[{Bai et~al.(2026)Bai, Wan, Zhou, Yu, You, and Tsang}]{bai2026skilldag}
Tong Bai, Zhenglin Wan, Pengfei Zhou, Xingrui Yu, Yang You, and Ivor~W. Tsang.
  2026.
\newblock \href {https://arxiv.org/abs/2606.03056} {{SkillDAG}: Self-evolving
  typed skill graphs for {LLM} skill selection at scale}.
\newblock \emph{Preprint}, arXiv:2606.03056.

\bibitem[{Chen et~al.(2026)Chen, Yu, Gu, Huang, Li, Liu, Liu, Liu, Peng, Wang,
  Yan, Meng, Qin, Che, and Hu}]{chen2026scaling}
Charles Chen, Qiming Yu, Yuhang Gu, Zhuoye Huang, Hanjing Li, Hongyu Liu, Simin
  Liu, Jinhao Liu, Dengyun Peng, Jiangyi Wang, Zheng Yan, Fanqing Meng, Ethan
  Qin, Carl Che, and Mengkang Hu. 2026.
\newblock \href {https://arxiv.org/abs/2605.16508} {The scaling laws of skills
  in {LLM} agent systems}.
\newblock \emph{Preprint}, arXiv:2605.16508.

\bibitem[{Cormack et~al.(2009)Cormack, Clarke, and Buettcher}]{cormack2009rrf}
Gordon~V. Cormack, Charles L.~A. Clarke, and Stefan Buettcher. 2009.
\newblock \href {https://doi.org/10.1145/1571941.1572114} {Reciprocal rank
  fusion outperforms {Condorcet} and individual rank learning methods}.
\newblock In \emph{Proceedings of the 32nd International ACM SIGIR Conference
  on Research and Development in Information Retrieval}, pages 758--759.

\bibitem[{De~Lange et~al.(2022)De~Lange, Aljundi, Masana, Parisot, Jia,
  Leonardis, Slabaugh, and Tuytelaars}]{delange2022survey}
Matthias De~Lange, Rahaf Aljundi, Marc Masana, Sarah Parisot, Xu~Jia,
  Ale{\v{s}} Leonardis, Gregory Slabaugh, and Tinne Tuytelaars. 2022.
\newblock \href {https://doi.org/10.1109/TPAMI.2021.3057446} {A continual
  learning survey: Defying forgetting in classification tasks}.
\newblock \emph{IEEE Transactions on Pattern Analysis and Machine
  Intelligence}, 44(7):3366--3385.

\bibitem[{{DeepSeek-AI}(2026)}]{deepseek2026v4}
{DeepSeek-AI}. 2026.
\newblock \href {https://arxiv.org/abs/2606.19348} {{DeepSeek-V4}: Towards
  highly efficient million-token context intelligence}.
\newblock \emph{Preprint}, arXiv:2606.19348.

\bibitem[{Ghoshal et~al.(2026)Ghoshal, Mittal, Singh, Ballesteros, Sun, Tu,
  Singh, Benajiba, Shah, Bharadwaj, Ravi, and Roth}]{ghoshal2026jtpro}
Sandip Ghoshal, Anshul Mittal, Jyotika Singh, Miguel Ballesteros, Weiyi Sun,
  Fang Tu, Shailender Singh, Yassine Benajiba, Fahad Shah, Sujeeth Bharadwaj,
  Sujith Ravi, and Dan Roth. 2026.
\newblock \href {https://doi.org/10.18653/v1/2026.findings-acl.2017} {{JTPRO}:
  A joint tool-prompt reflective optimization framework for language agents}.
\newblock In \emph{Findings of the Association for Computational Linguistics:
  ACL 2026}, pages 40573--40595.

\bibitem[{Gillespie and Perry(2026)}]{gillespie2026enterprise}
Kellen Gillespie and Robyn Perry. 2026.
\newblock \href {https://arxiv.org/abs/2606.17519} {Scaling enterprise agent
  routing: Degradation, diagnosis, and recovery}.
\newblock \emph{Preprint}, arXiv:2606.17519.

\bibitem[{{Google DeepMind}(2026{\natexlab{a}})}]{google2026gemini35flashlite}
{Google DeepMind}. 2026{\natexlab{a}}.
\newblock \href
  {https://deepmind.google/models/model-cards/gemini-3-5-flash-lite/} {{Gemini}
  3.5 {Flash-Lite} model card}.

\bibitem[{{Google DeepMind}(2026{\natexlab{b}})}]{google2026gemini36flash}
{Google DeepMind}. 2026{\natexlab{b}}.
\newblock \href {https://deepmind.google/models/model-cards/gemini-3-6-flash/}
  {{Gemini} 3.6 {Flash} model card}.

\bibitem[{Guo et~al.(2026)Guo, Dong, Gao, and Das}]{guo2026tracefree}
Ruocheng Guo, Kaiwen Dong, Xiang Gao, and Kamalika Das. 2026.
\newblock \href {https://arxiv.org/abs/2602.20426} {Learning to rewrite tool
  descriptions for reliable {LLM}-agent tool use}.
\newblock \emph{Preprint}, arXiv:2602.20426.

\bibitem[{Hao et~al.(2026)Hao, Min, Zhang, Zhang, Xu, Ge, Yu, and
  Cheng}]{hao2026skillinjection}
Haochang Hao, Dehai Min, Zhifang Zhang, Yunbei Zhang, Miao Xu, Yingqiang Ge,
  Philip~S. Yu, and Lu~Cheng. 2026.
\newblock \href {https://arxiv.org/abs/2606.07943} {One skill, two masters:
  Serving users and attackers through skill injection}.
\newblock \emph{Preprint}, arXiv:2606.07943.

\bibitem[{Hu et~al.(2026)Hu, Long, and Wang}]{hu2026memorycl}
Qisheng Hu, Quanyu Long, and Wenya Wang. 2026.
\newblock \href {https://arxiv.org/abs/2604.27003} {When continual learning
  moves to memory: A study of experience reuse in {LLM} agents}.
\newblock \emph{Preprint}, arXiv:2604.27003.

\bibitem[{Huang et~al.(2026)Huang, Zhang, Liang, Bei, Chen, Feng, Pan, Tan,
  Wang, Wei, Wu, Xu, Yang, Yang, Yang, Yeh, Zhang, Zhang, Zhu, Zou, Zhao, Wang,
  Xu, Ke, Hui, Li, Wu, He, Wang, Xu, Huang, Tan, Heinecke, Wang, Xiong,
  Metwally, Yan, Lee, Zeng, Xia, Wei, Payani, Wang, Ma, Wang, Wang, Zhang,
  Wang, Zhang, You, Tong, Luo, Liu, Sun, Wang, McAuley, Zou, Han, Yu, and
  Shu}]{huang2026memorysurvey}
Wei-Chieh Huang, Weizhi Zhang, Yueqing Liang, Yuanchen Bei, Yankai Chen, Tao
  Feng, Xinyu Pan, Zhen Tan, Yu~Wang, Tianxin Wei, Shanglin Wu, Ruiyao Xu,
  Liangwei Yang, Rui Yang, Wooseong Yang, Chin-Yuan Yeh, Hanrong Zhang, Haozhen
  Zhang, Siqi Zhu, and 41 others. 2026.
\newblock \href {https://arxiv.org/abs/2602.06052} {A survey of agent memory in
  the second half: Towards self-evolving and long-horizon agents}.
\newblock \emph{Transactions on Machine Learning Research}.

\bibitem[{Kang(2026)}]{kang2026skillseam}
Ruiyuan Kang. 2026.
\newblock \href {https://arxiv.org/abs/2609.13321} {{SkillSeam}: Six principles
  for auditing agent skill collections}.
\newblock \emph{Preprint}, arXiv:2609.13321.

\bibitem[{Kang et~al.(2026)Kang, Cho, and Kim}]{kang2026skillret}
Ryangkyung Kang, Hongcheol Cho, and Youngeun Kim. 2026.
\newblock \href {https://arxiv.org/abs/2605.05726} {{SkillRet}: A large-scale
  benchmark for skill retrieval in {LLM} agents}.
\newblock \emph{Preprint}, arXiv:2605.05726.

\bibitem[{Kirkpatrick et~al.(2017)Kirkpatrick, Pascanu, Rabinowitz, Veness,
  Desjardins, Rusu, Milan, Quan, Ramalho, Grabska-Barwinska, Hassabis, Clopath,
  Kumaran, and Hadsell}]{kirkpatrick2017overcoming}
James Kirkpatrick, Razvan Pascanu, Neil Rabinowitz, Joel Veness, Guillaume
  Desjardins, Andrei~A. Rusu, Kieran Milan, John Quan, Tiago Ramalho, Agnieszka
  Grabska-Barwinska, Demis Hassabis, Claudia Clopath, Dharshan Kumaran, and
  Raia Hadsell. 2017.
\newblock \href {https://doi.org/10.1073/pnas.1611835114} {Overcoming
  catastrophic forgetting in neural networks}.
\newblock \emph{Proceedings of the National Academy of Sciences},
  114(13):3521--3526.

\bibitem[{Koehn(2004)}]{koehn2004significance}
Philipp Koehn. 2004.
\newblock \href {https://aclanthology.org/W04-3250/} {Statistical significance
  tests for machine translation evaluation}.
\newblock In \emph{Proceedings of the 2004 Conference on Empirical Methods in
  Natural Language Processing}, pages 388--395.

\bibitem[{Liu et~al.(2026)Liu, Ji, An, Jaakkola, Zhang, and
  Chang}]{liu2026skillswild}
Yujian Liu, Jiabao Ji, Li~An, Tommi Jaakkola, Yang Zhang, and Shiyu Chang.
  2026.
\newblock \href {https://arxiv.org/abs/2604.04323} {How well do agentic skills
  work in the wild: Benchmarking {LLM} skill usage in realistic settings}.
\newblock \emph{Preprint}, arXiv:2604.04323.

\bibitem[{Lopez-Paz and Ranzato(2017)}]{lopezpaz2017gem}
David Lopez-Paz and Marc'Aurelio Ranzato. 2017.
\newblock \href
  {https://proceedings.neurips.cc/paper/2017/hash/f87522788a2be2d171666752f97ddebb-Abstract.html}
  {Gradient episodic memory for continual learning}.
\newblock In \emph{Advances in Neural Information Processing Systems},
  volume~30, pages 6467--6476.

\bibitem[{Lu et~al.(2026{\natexlab{a}})Lu, Huang, Meng, Jin, Zeng, and
  Shen}]{lu2025toolde}
Xuan Lu, Haohang Huang, Rui Meng, Yaohui Jin, Wenjun Zeng, and Xiaoyu Shen.
  2026{\natexlab{a}}.
\newblock \href {https://openreview.net/forum?id=g9D9MgG7iW} {Tools are
  under-documented: Simple document expansion boosts tool retrieval}.
\newblock In \emph{The Fourteenth International Conference on Learning
  Representations}.

\bibitem[{Lu et~al.(2026{\natexlab{b}})Lu, Zhang, Chen, and
  Peng}]{lu2026docschisel}
You Lu, Kun Zhang, Bihuan Chen, and Xin Peng. 2026{\natexlab{b}}.
\newblock \href {https://arxiv.org/abs/2608.10037} {{DocsChisel}: Adaptive tool
  documentation optimization framework for {LLM} agents}.
\newblock \emph{Preprint}, arXiv:2608.10037.

\bibitem[{Murtaza et~al.(2026)Murtaza, Nie, Soni, Wen, and
  Frydenlund}]{murtaza2026synthetic}
Syed~Shariyar Murtaza, Yifan Nie, Utkarsh Soni, Eugene Wen, and Arvid
  Frydenlund. 2026.
\newblock \href {https://arxiv.org/abs/2609.10750} {When synthetic data hurts:
  On catastrophic forgetting in skill retrieval for {LLM} agents}.
\newblock \emph{Preprint}, arXiv:2609.10750.

\bibitem[{{OpenAI}(2025)}]{openai2025codexcli}
{OpenAI}. 2025.
\newblock \href {https://github.com/openai/codex} {Codex {CLI}: A lightweight
  coding agent that runs in your terminal}.

\bibitem[{{OpenAI}(2026{\natexlab{a}})}]{openai2026gpt55}
{OpenAI}. 2026{\natexlab{a}}.
\newblock \href {https://openai.com/index/gpt-5-5-system-card} {{GPT-5.5}
  system card}.

\bibitem[{{OpenAI}(2026{\natexlab{b}})}]{openai2026gpt56}
{OpenAI}. 2026{\natexlab{b}}.
\newblock \href {https://deploymentsafety.openai.com/gpt-5-6} {{GPT-5.6} system
  card}.

\bibitem[{{OpenAI}(2026{\natexlab{c}})}]{openai2026gpt54mini}
{OpenAI}. 2026{\natexlab{c}}.
\newblock \href {https://openai.com/index/introducing-gpt-5-4-mini-and-nano/}
  {Introducing {GPT-5.4} mini and nano}.

\bibitem[{{Qwen Team}(2026)}]{qwen2026qwen37flash}
{Qwen Team}. 2026.
\newblock {Qwen3.7-Flash}.
\newblock QwenCloud model release, 25 July 2026.

\bibitem[{Reimers and Gurevych(2019)}]{reimers2019sbert}
Nils Reimers and Iryna Gurevych. 2019.
\newblock \href {https://doi.org/10.18653/v1/D19-1410} {{Sentence-BERT}:
  Sentence embeddings using {Siamese} {BERT}-networks}.
\newblock In \emph{Proceedings of the 2019 Conference on Empirical Methods in
  Natural Language Processing and the 9th International Joint Conference on
  Natural Language Processing (EMNLP-IJCNLP)}, pages 3982--3992.

\bibitem[{Robertson and Zaragoza(2009)}]{robertson2009bm25}
Stephen Robertson and Hugo Zaragoza. 2009.
\newblock \href {https://doi.org/10.1561/1500000019} {The probabilistic
  relevance framework: {BM25} and beyond}.
\newblock \emph{Foundations and Trends in Information Retrieval},
  3(4):333--389.

\bibitem[{Song et~al.(2026)Song, Zhang, Liu, Liu, Li, Yuan, Zhang, Zou, Yan,
  Zhang, Ye, Yu, and Sun}]{song2026drclaw}
Dingjie Song, Hanrong Zhang, Dawei Liu, Yixin Liu, Zongxia Li, Zhengqing Yuan,
  Siqi Zhang, Henry~Peng Zou, Zhiling Yan, Yuxuan Zhang, Yanfang Ye, Philip~S.
  Yu, and Lichao Sun. 2026.
\newblock \href {https://arxiv.org/abs/2609.00365} {{Dr. Claw}: An {AI}
  scientist workspace for vibe research}.
\newblock In \emph{Proceedings of the 2026 Conference on Empirical Methods in
  Natural Language Processing: System Demonstrations}.

\bibitem[{Song and Wei(2026)}]{song2026shadowing}
Hongwen Song and Song Wei. 2026.
\newblock \href {https://arxiv.org/abs/2605.24050} {More skills, worse agents?
  {Skill} shadowing degrades performance when expanding skill libraries}.
\newblock \emph{Preprint}, arXiv:2605.24050.

\bibitem[{Su et~al.(2026)Su, Long, Ai, He, Tang, Wang, Tu, Wang, and
  Liu}]{su2026sra}
Weihang Su, Jianming Long, Qingyao Ai, Qiaozhi He, Yichen Tang, Changyue Wang,
  Yiteng Tu, Yingbo Wang, and Yiqun Liu. 2026.
\newblock \href {https://arxiv.org/abs/2604.24594} {Skill retrieval
  augmentation for agentic ai}.
\newblock \emph{Preprint}, arXiv:2604.24594.
\newblock Author repository: \url{https://github.com/oneal2000/SR-Agents}.

\bibitem[{van~de Ven et~al.(2022)van~de Ven, Tuytelaars, and
  Tolias}]{vandeven2022three}
Gido~M. van~de Ven, Tinne Tuytelaars, and Andreas~S. Tolias. 2022.
\newblock \href {https://doi.org/10.1038/s42256-022-00568-3} {Three types of
  incremental learning}.
\newblock \emph{Nature Machine Intelligence}, 4(12):1185--1197.

\bibitem[{van~den Oord et~al.(2018)van~den Oord, Li, and Vinyals}]{oord2018cpc}
Aaron van~den Oord, Yazhe Li, and Oriol Vinyals. 2018.
\newblock \href {https://arxiv.org/abs/1807.03748} {Representation learning
  with contrastive predictive coding}.
\newblock \emph{Preprint}, arXiv:1807.03748.

\bibitem[{Wang et~al.(2024)Wang, Xie, Jiang, Mandlekar, Xiao, Zhu, Fan, and
  Anandkumar}]{wang2024voyager}
Guanzhi Wang, Yuqi Xie, Yunfan Jiang, Ajay Mandlekar, Chaowei Xiao, Yuke Zhu,
  Linxi Fan, and Anima Anandkumar. 2024.
\newblock \href {https://arxiv.org/abs/2305.16291} {Voyager: An open-ended
  embodied agent with large language models}.
\newblock \emph{Transactions on Machine Learning Research}.

\bibitem[{Wang et~al.(2020)Wang, Wei, Dong, Bao, Yang, and
  Zhou}]{wang2020minilm}
Wenhui Wang, Furu Wei, Li~Dong, Hangbo Bao, Nan Yang, and Ming Zhou. 2020.
\newblock {MiniLM}: Deep self-attention distillation for task-agnostic
  compression of pre-trained transformers.
\newblock In \emph{Advances in Neural Information Processing Systems},
  volume~33, pages 5776--5788.

\bibitem[{Wang et~al.(2026{\natexlab{a}})Wang, Wen, Ji, Chen, Qiao, and
  Sun}]{wang2026r3skill}
Zifei Wang, Wei Wen, Qiang Ji, Keyu Chen, Ruizhi Qiao, and Xing Sun.
  2026{\natexlab{a}}.
\newblock \href {https://arxiv.org/abs/2606.03565} {Skill is not document:
  Query-conditioned compatibility for {LLM} agent skill routing}.
\newblock \emph{Preprint}, arXiv:2606.03565.

\bibitem[{Wang et~al.(2026{\natexlab{b}})Wang, Wen, Ji, and
  Qiao}]{wang2026capability}
Zifei Wang, Wei Wen, Qiang Ji, and Ruizhi Qiao. 2026{\natexlab{b}}.
\newblock \href {https://arxiv.org/abs/2608.04482} {Skills know their
  neighbors: Cluster-contrastive capability pages for skill retrieval}.
\newblock \emph{Preprint}, arXiv:2608.04482.

\bibitem[{Wu et~al.(2026)Wu, Zhang, Lin, Xu, Xu, Chen, Zou, Chen, Zhang, Liu,
  Yu, and Wang}]{wu2026gam}
Zhaofen Wu, Hanrong Zhang, Fulin Lin, Wujiang Xu, Xinran Xu, Yankai Chen,
  Henry~Peng Zou, Shaowen Chen, Weizhi Zhang, Xue Liu, Philip~S. Yu, and
  Hongwei Wang. 2026.
\newblock \href {https://doi.org/10.18653/v1/2026.acl-long.1600} {{GAM}:
  Hierarchical graph-based agentic memory for {LLM} agents}.
\newblock In \emph{Proceedings of the 64th Annual Meeting of the Association
  for Computational Linguistics (Volume 1: Long Papers)}, pages 34647--34664.

\bibitem[{Xu et~al.(2025)Xu, Liang, Mei, Gao, Tan, and Zhang}]{xu2025amem}
Wujiang Xu, Zujie Liang, Kai Mei, Hang Gao, Juntao Tan, and Yongfeng Zhang.
  2025.
\newblock \href {https://doi.org/10.52202/085713-0593} {{A-Mem}: Agentic memory
  for {LLM} agents}.
\newblock In \emph{Advances in Neural Information Processing Systems},
  volume~38, pages 17577--17604.

\bibitem[{Yan et~al.(2026)Yan, Song, Zhang, Liang, Zhang, Dai, He, Yu, Xu, Li,
  and Sun}]{yan2026openskill}
Zhiling Yan, Dingjie Song, Hanrong Zhang, Wei Liang, Yuxuan Zhang, Yutong Dai,
  Lifang He, Philip~S. Yu, Ran Xu, Xiang Li, and Lichao Sun. 2026.
\newblock \href {https://arxiv.org/abs/2606.06741} {{OpenSkill}: Open-world
  self-evolution for {LLM} agents}.
\newblock In \emph{Proceedings of the 2026 Conference on Empirical Methods in
  Natural Language Processing (EMNLP)}.

\bibitem[{Yang et~al.(2026)Yang, Zheng, Sun, Fan, Wang, Wang, Xu, Liu, Fang,
  Cai, and Yu}]{yang2026timeevo}
Jie Yang, Yan Zheng, Jiarui Sun, Xiran Fan, Junpeng Wang, Liang Wang, Zelin Xu,
  Qinghua Liu, Zhengyu Fang, Yiwei Cai, and Philip~S. Yu. 2026.
\newblock \href {https://arxiv.org/abs/2609.27277} {{TimeEvo}: Failure-driven
  self-evolution of a time series agent}.
\newblock \emph{Preprint}, arXiv:2609.27277.

\bibitem[{Yao et~al.(2026)Yao, Zhang, Lin, Jin, Li, Wang, and
  Chi}]{yao2026uncertainty}
Yifei Yao, Hanrong Zhang, Fulin Lin, Ziyan Jin, Xiaoxiao Li, Hongwei Wang, and
  Ying Chi. 2026.
\newblock \href {https://doi.org/10.1109/JBHI.2026.3701922} {From uncertainty
  to clarity: Uncertainty-guided class-incremental learning for limited
  biomedical samples via semantic expansion}.
\newblock \emph{IEEE Journal of Biomedical and Health Informatics}, pages
  1--14.

\bibitem[{Zhang et~al.(2026{\natexlab{a}})Zhang, Liu, Chen, and
  Ji}]{zhang2026skillmd}
Chi Zhang, Yimin Liu, Xinze Chen, and Ping Ji. 2026{\natexlab{a}}.
\newblock \href {https://arxiv.org/abs/2608.08453} {What keeps agent skills
  from being reusable? evidence from {138K} {SKILL.md} files}.
\newblock \emph{Preprint}, arXiv:2608.08453.

\bibitem[{Zhang et~al.(2026{\natexlab{b}})Zhang, Fan, Zou, Chen, Wang, Zhou,
  Li, Huang, Yao, Zheng, Liu, Li, and Yu}]{zhang2026coevoskills}
Hanrong Zhang, Shicheng Fan, Henry~Peng Zou, Yankai Chen, Zhenting Wang, Jiayu
  Zhou, Chengze Li, Wei-Chieh Huang, Yifei Yao, Kening Zheng, Xue Liu, Xiaoxiao
  Li, and Philip~S. Yu. 2026{\natexlab{b}}.
\newblock \href {https://arxiv.org/abs/2604.01687} {{CoEvoSkills}:
  Self-evolving agent skills via co-evolutionary verification}.
\newblock In \emph{Conference on Language Modeling (COLM)}.

\bibitem[{Zhang et~al.(2025{\natexlab{a}})Zhang, Huang, Mei, Yao, Wang, Zhan,
  Wang, and Zhang}]{zhang2025asb}
Hanrong Zhang, Jingyuan Huang, Kai Mei, Yifei Yao, Zhenting Wang, Chenlu Zhan,
  Hongwei Wang, and Yongfeng Zhang. 2025{\natexlab{a}}.
\newblock \href {https://openreview.net/forum?id=V4y0CpX4hK} {Agent security
  bench ({ASB}): Formalizing and benchmarking attacks and defenses in
  {LLM}-based agents}.
\newblock In \emph{International Conference on Learning Representations
  (ICLR)}.

\bibitem[{Zhang et~al.(2025{\natexlab{b}})Zhang, Yao, Wang, Su, Li, Peng, and
  Wang}]{zhang2025classincremental}
Hanrong Zhang, Yifei Yao, Zixuan Wang, Jiayuan Su, Mengxuan Li, Peng Peng, and
  Hongwei Wang. 2025{\natexlab{b}}.
\newblock \href {https://doi.org/10.1109/TII.2025.3534407} {Class incremental
  fault diagnosis under limited fault data via supervised contrastive knowledge
  distillation}.
\newblock \emph{IEEE Transactions on Industrial Informatics}, 21(6):4344--4354.

\bibitem[{Zhang et~al.(2025{\natexlab{c}})Zhang, Li, Long, Zhang, Lin, Yang,
  Xie, Yang, Liu, Lin, Huang, and Zhou}]{zhang2025qwen3embedding}
Yanzhao Zhang, Mingxin Li, Dingkun Long, Xin Zhang, Huan Lin, Baosong Yang,
  Pengjun Xie, An~Yang, Dayiheng Liu, Junyang Lin, Fei Huang, and Jingren Zhou.
  2025{\natexlab{c}}.
\newblock \href {https://arxiv.org/abs/2506.05176} {{Qwen3 Embedding}:
  Advancing text embedding and reranking through foundation models}.
\newblock \emph{Preprint}, arXiv:2506.05176.

\bibitem[{Zheng et~al.(2026)Zheng, Zhang, Ma, Yu, Zhu, Wu, Xu, Dong, Zhu,
  Huang, and Yu}]{zheng2026skillrouter}
YanZhao Zheng, ZhenTao Zhang, Chao Ma, YuanQiang Yu, JiHuai Zhu, Yong Wu,
  Tianze Xu, Baohua Dong, Hangcheng Zhu, Ruohui Huang, and Gang Yu. 2026.
\newblock \href {https://arxiv.org/abs/2603.22455} {{SkillRouter}: Skill
  routing for {LLM} agents at scale}.
\newblock \emph{Preprint}, arXiv:2603.22455.

\bibitem[{Zhou et~al.(2024)Zhou, Wang, Qi, Ye, Zhan, and
  Liu}]{zhou2024classincremental}
Da-Wei Zhou, Qi-Wei Wang, Zhi-Hong Qi, Han-Jia Ye, De-Chuan Zhan, and Ziwei
  Liu. 2024.
\newblock \href {https://doi.org/10.1109/TPAMI.2024.3429383} {Class-incremental
  learning: A survey}.
\newblock \emph{IEEE Transactions on Pattern Analysis and Machine
  Intelligence}, 46(12):9851--9873.

\bibitem[{Zhou et~al.(2026)Zhou, Alqudah, Lai, Halfaker, Xiong, and
  Harari}]{zhou2026rewrite}
Yangqiaoyu Zhou, Mohammad Alqudah, Kwei-Herng Lai, Aaron Halfaker, Yingqi
  Xiong, and Yaar Harari. 2026.
\newblock \href {https://arxiv.org/abs/2606.30775} {A single rewrite suffices:
  Empirical lessons from production skill description optimization}.
\newblock \emph{Preprint}, arXiv:2606.30775.

\bibitem[{Zhu et~al.(2026)Zhu, Fan, Wang, Wu, Zhou, and
  Huang}]{zhu2026rsiagent}
Sibo Zhu, Shicheng Fan, Xinyue Wang, Wenyi Wu, Kun Zhou, and Biwei Huang. 2026.
\newblock \href {https://arxiv.org/abs/2609.15364} {{RSIAgent}: Autonomous
  exploration for recursive self-improvement in new environments}.
\newblock \emph{Preprint}, arXiv:2609.15364.

\bibitem[{Zhuo et~al.(2025)Zhuo, Vu, Chim, Hu, Yu, Widyasari, Yusuf, Zhan, He,
  Paul, Brunner, Gong, Hoang, Zebaze, Hong, Li, Kaddour, Xu, Zhang, Yadav,
  Jain, Gu, Cheng, Liu, Liu, Wang, Hui, Muennighoff, Lo, Fried, Du, de~Vries,
  and von Werra}]{zhuo2025bigcodebench}
Terry~Yue Zhuo, Minh~Chien Vu, Jenny Chim, Han Hu, Wenhao Yu, Ratnadira
  Widyasari, Imam Nur~Bani Yusuf, Haolan Zhan, Junda He, Indraneil Paul, Simon
  Brunner, Chen Gong, Thong Hoang, Armel~Randy Zebaze, Xiaoheng Hong, Wen-Ding
  Li, Jean Kaddour, Ming Xu, Zhihan Zhang, and 14 others. 2025.
\newblock \href {https://arxiv.org/abs/2406.15877} {{BigCodeBench}:
  Benchmarking code generation with diverse function calls and complex
  instructions}.
\newblock In \emph{Proceedings of the Thirteenth International Conference on
  Learning Representations}.

\bibitem[{Zou et~al.(2026{\natexlab{a}})Zou, Huang, Wu, Guo, Chen, Miao,
  Nguyen, Zhou, Zhang, Fang, Zhang, Wang, Zhang, He, Li, Li, Jiang, and
  Yu}]{zou2026humanagent}
Henry~Peng Zou, Wei-Chieh Huang, Yaozu Wu, Jizhou Guo, Yankai Chen, Chunyu
  Miao, Hoang~H. Nguyen, Yue Zhou, Weizhi Zhang, Liancheng Fang, Hanrong Zhang,
  Fangxin Wang, Pengfei Zhang, Langzhou He, Yangning Li, Dongyuan Li, Renhe
  Jiang, and Philip~S. Yu. 2026{\natexlab{a}}.
\newblock \href {https://doi.org/10.18653/v1/2026.findings-acl.1811}
  {{LLM}-based human-agent collaboration and interaction systems: A survey}.
\newblock In \emph{Findings of the Association for Computational Linguistics:
  ACL 2026}, pages 36335--36364.

\bibitem[{Zou et~al.(2026{\natexlab{b}})Zou, Miao, Huang, Chen, Zhou, Zhang,
  Wu, Fang, Gu, Zhang, Zheng, Wang, Nian, Li, Fan, He, Zhang, Liu, and
  Yu}]{zou2026interruptible}
Henry~Peng Zou, Chunyu Miao, Wei-Chieh Huang, Yankai Chen, Yue Zhou, Hanrong
  Zhang, Yaozu Wu, Liancheng Fang, Zhengyao Gu, Zhen Zhang, Kening Zheng,
  Fangxin Wang, Yi~Nian, Shanghao Li, Wenzhe Fan, Langzhou He, Weizhi Zhang,
  Xue Liu, and Philip~S. Yu. 2026{\natexlab{b}}.
\newblock \href {https://arxiv.org/abs/2604.00892} {When users change their
  mind: Evaluating interruptible agents in long-horizon web navigation}.
\newblock \emph{Preprint}, arXiv:2604.00892.

\end{thebibliography}
